%% file: iclr2027_conference.tex
\documentclass{article} % For LaTeX2e
\usepackage{iclr2027_conference,times}

\input{math_commands.tex}

\usepackage{hyperref}
\usepackage{url}

\usepackage{hyperref}
\renewcommand{\arraystretch}{0.2}  % 调整行距
\usepackage{comment}
\usepackage{amsmath}
\usepackage{gensymb}
\usepackage{amssymb}
\usepackage{times}
\usepackage{booktabs}
\usepackage{array}
\usepackage{makecell}
\usepackage{pifont}
\usepackage{caption}
\usepackage[T1]{fontenc}
\usepackage{multirow}
\usepackage{hyperref}
\usepackage{bm}
\usepackage{graphicx}
\usepackage{adjustbox}
\usepackage{longtable}    % 支持跨页表格
\usepackage{tabularx}
\usepackage{array}    % for column definitions
\usepackage{comment}
\usepackage[table]{xcolor}
\usepackage{tabularx}
\usepackage{algorithm,algorithmic}

\usepackage{makecell}
\usepackage{amsthm}
\usepackage[english]{babel}
\theoremstyle{plain}
\newtheorem{theorem}{Theorem}

\theoremstyle{definition}

\theoremstyle{remark}

\usepackage{enumitem}

\usepackage{wrapfig}
\usepackage{algorithm}
\usepackage{algorithmic}
\usepackage{amsmath,amssymb,bm}

\usepackage{tikz}

\UseRawInputEncoding

\usepackage{makecell}
\title{Learning Intrinsic Water-Quality Dynamics with Rainfall for Data-Driven Forecasting}
\author{
Ziqi Wang, Hailiang Zhao, Cheng Bao, Daojiang Hu, Wenzhuo Qian, Shuiguang Deng
\\[8pt]
\makebox[\linewidth][c]{\normalfont Zhejiang University}
}

\iclrfinalcopy % Uncomment for camera-ready version, but NOT for submission.
\begin{document}

\maketitle
\fancyhead{}
\begin{abstract}
% 降雨是影响水质变化的重要外部因素 → 传统上常通过机理模型描述这种作用，但依赖过程假设、参数标定和场景适配 → 本文尝试用数据驱动方式学习降雨与水质之间复杂、动态的关系。
% Rainfall is an important environmental driver of water-quality variations, affecting hydrological processes such as runoff, pollutant transport, dilution, and resuspension. Traditional mechanistic models can explicitly describe these processes, but often rely on substantial process specification, site-specific calibration, and relatively stable environmental relationships, limiting their flexibility under changing hydrological conditions. In this work, we explore a data-driven alternative by proposing RaiNet to jointly model multiscale water-quality dynamics and station-specific rainfall effects across relative lags and temporal scales. It models water-quality dynamics with \texttt{LocTrend}, constructs station-oriented rainfall events from gridded precipitation, and uses \texttt{XGateFusion} for selective lag-aware fusion. We further release three real-world multimodal datasets, each containing two years of hourly water-quality observations aligned with gridded rainfall data. Experiments show that RaiNet consistently outperforms competitive time-series, spatiotemporal, multimodal fusion, and large time-series forecasting baselines, while component-wise analyses further demonstrate the distinct roles and effectiveness of each module.
Rainfall is an important environmental driver of water-quality variations through processes such as runoff, pollutant transport, dilution, and resuspension. Traditional mechanistic models can explicitly describe these processes but often require substantial process specification and site-specific calibration, limiting their flexibility under changing hydrological conditions. In this work, we explore a data-driven alternative by proposing RaiNet to jointly model multiscale water-quality dynamics and station-specific rainfall effects across relative lags and temporal scales. RaiNet employs \texttt{LocTrend} to capture irregular water-quality dynamics, constructs station-oriented rainfall events from gridded precipitation, and introduces \texttt{XGateFusion} for conditional lag-aware fusion across scales. We further release three real-world multimodal datasets comprising over 150,000 temporally aligned water quality observations and gridded precipitation raster images. Experiments show that RaiNet outperforms general time-series, water quality, diffusion-based, and spatiotemporal models by over 20\%, while component-wise analyses confirm the distinct contribution of each module.

%We further release three real-world multimodal datasets, each comprising two years of hourly water-quality observations aligned with gridded rainfall data. Experiments demonstrate that RaiNet consistently outperforms strong time-series, spatiotemporal, and models, while component-wise analyses verify the distinct contribution of each module.

\end{abstract}

% --------------------------------------------------------------------------
\section{Introduction}\label{sec:introduction}
Accurate water-quality forecasting is essential for environmental management and ecosystem protection, but remains challenging because hydrological, meteorological, and episodic influences vary over time and across stations. Mechanistic models such as SWAT \citep{krysanova2015advances}, WASP \citep{wool2020wasp}, and QUAL2K \citep{zhang2012simulation} provide interpretable predictions but require detailed process specification, extensive environmental inputs, and site-specific calibration, limiting their transferability under changing catchment conditions. These limitations are particularly restrictive when the effects of rainfall vary over space and time. This motivates more flexible data-driven approaches that can learn evolving environmental dependencies directly from observations.
%Accurate water-quality forecasting is essential for environmental management and ecosystem protection, but remains challenging because hydrological, meteorological, and episodic influences vary over time and across stations. Mechanistic models such as SWAT \citep{krysanova2015advances}, WASP \citep{wool2020wasp}, and QUAL2K \citep{zhang2012simulation} provide physically interpretable predictions but require detailed process specification, extensive environmental inputs, and site-specific calibration, limiting their transferability under changing catchment conditions. These limitations are restrictive when water-quality responses are driven by heterogeneous external forcing, such as rainfall, whose effects vary across locations and temporal scales. This motivates more flexible data-driven approaches that learn evolving environmental dependencies directly from observations.

\begin{figure}[!htbp]
    \centering
    \includegraphics[width=1\linewidth]{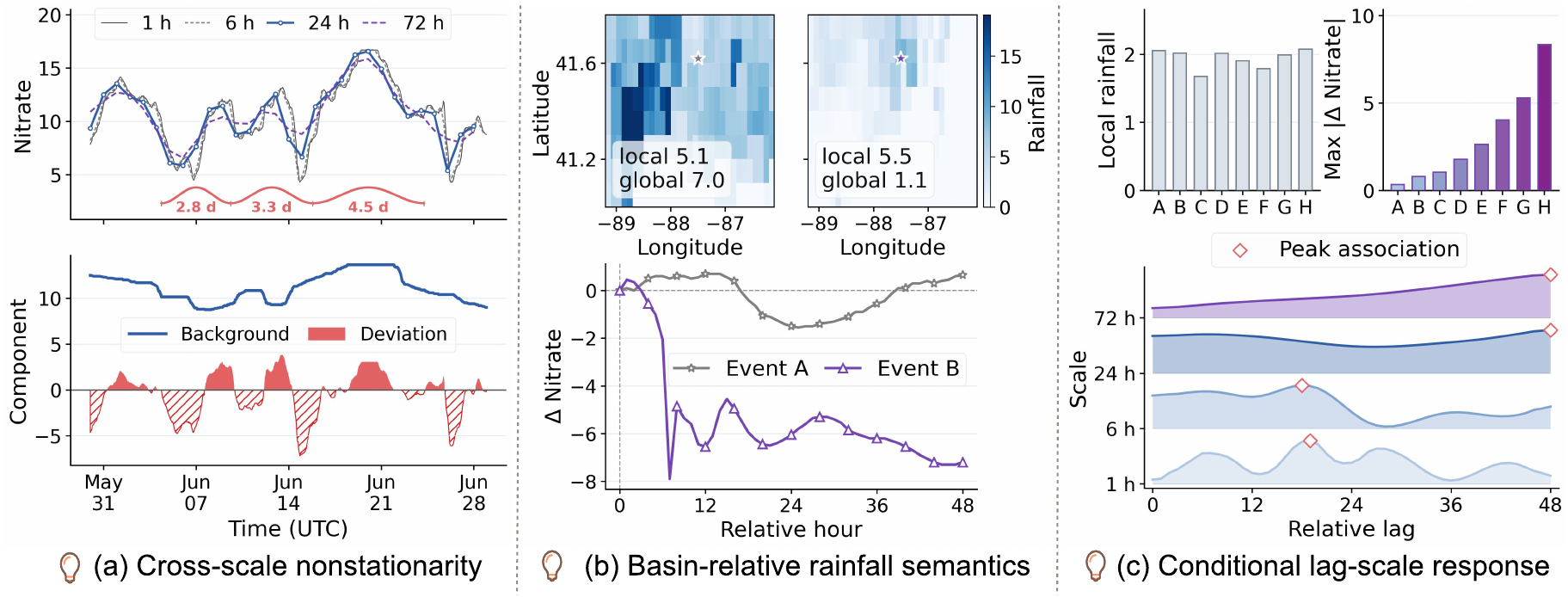}
    \caption{\textbf{Empirical motivations of this work}. (a) Water-quality dynamics vary across temporal resolutions, with irregular background-deviation patterns and no fixed periodicity.
(b) Similar local rainfall can represent different events under different basin-wide contexts, leading to heterogeneous water-quality variations.
(c) Comparable rainfall events can induce different water-quality responses, with strongest associations appearing at different relative lags across temporal scales.}
% (a) Water-quality dynamics vary across temporal resolutions and exhibit irregular background-deviation patterns without a fixed periodicity.
% (b) Similar station-local rainfall can represent different events under different basin-wide rainfall backgrounds and is followed by heterogeneous water-quality variations.
% (c) Comparable rainfall events can produce substantially different water-quality responses, while the strongest rainfall-water-quality associations occur at different relative lags across temporal scales.}
    \label{fig1}
    \vspace{-1.2em}
\end{figure}

Recent time-series and spatiotemporal models can learn complex nonlinear dependencies from observations \citep{TimesNet,FEDformer,graph111,VT,TimeKAN}. However, general-purpose forecasting architectures often overlook the influence of external environmental conditions on water-quality dynamics. We focus on a specific yet challenging factor: \textit{rainfall}. Its effects on water quality vary across stations through processes such as runoff, pollutant transport, and dilution. With increasingly available gridded precipitation products, rainfall can be integrated with water-quality observations as an additional modality for multimodal forecasting.

Effectively integrating the two modalities remains challenging because rainfall influence is intermittent, station-specific, lagged, and scale dependent. Existing multimodal fusion methods rarely model these conditional dependencies explicitly, limiting their ability to determine \textit{when, at what lag, and at which temporal scale} rainfall should contribute to water-quality forecasting \citep{CDA,MBT}. As illustrated in Figure~\ref{fig1}, this leads to three challenges. First, water-quality dynamics are multiscale and nonstationary, with short-lived fluctuations superimposed on slowly evolving backgrounds and no fixed periodic structure. Second, raw rainfall intensity does not directly represent a station-relevant event: similar local rainfall can have different meanings under different basin-wide contexts, while transient rainfall may not persist long enough to affect water quality. Third, rainfall and water-quality relationships are conditional, lagged, and scale dependent: similar rainfall events may correspond to different subsequent responses, and their strongest statistical associations can occur at different relative temporal offsets across scales. 

To address these challenges, we propose \textbf{RaiNet}, a rainfall-informed water-quality forecasting network. It first models intrinsic multiscale water-quality dynamics, constructs station-oriented rainfall events from gridded precipitation, and then selectively uses rainfall to correct independently predicted water-quality baselines. Our main contributions are:
\vspace{-0.5em}
\begin{itemize}
\item We develop a multiscale background-deviation modeling mechanism for irregular water-quality dynamics. Multiscale downsampling exposes temporal structures at different resolutions, while \texttt{LocTrend} adaptively separates locally evolving structural backgrounds from dynamic deviations without relying on predefined temporal patterns.
\item We introduce a rainfall-event encoder. By combining basin-to-local spatial contrast with temporal persistence, it transforms raw rainfall grids into station-aligned event representations that better characterize environmentally meaningful rainfall conditions.
\item We propose \texttt{XGateFusion}, a conditional, lag-aware, and scale-dependent fusion mechanism. It learns statistical rainfall and water-quality relationships across relative temporal offsets and adaptively determines whether and to what extent rainfall should correct the independently predicted water-quality baseline at each temporal scale. 
\end{itemize}
\vspace{-0.5em}
We establish three multimodal benchmarks with over 150,000 aligned water quality observations and gridded precipitation records, supporting research on multimodal environmental modeling.

%Experiments against general time-series models, spatiotemporal forecasting models, multimodal fusion methods, and large time-series models show that RaiNet consistently improves forecasting accuracy, while ablation and diagnostic analyses validate its components.

%时空效果在这个问题不适用
\section{Related Works}\label{sec:related}

\paragraph{Water-Quality Forecasting.} Existing water-quality forecasting methods broadly include mechanistic, statistical, and deep learning approaches. Mechanistic models, such as SWAT \citep{krysanova2015advances}, WASP \citep{wool2020wasp}, and QUAL2K \citep{zhang2012simulation}, explicitly represent hydrological and biogeochemical processes but require detailed inputs, process assumptions, and site-specific calibration. Statistical methods simplify this process by extracting temporal regularities through autoregression and seasonality, including ARIMA \citep{ARIMAnew}, SARIMA \citep{SARIMA}, Holt-Winters \citep{HW}, and STL \citep{STL}. General-purpose deep time-series models extend this data-driven direction to nonlinear dynamics: TimesNet \citep{TimesNet} models period-dependent variations, FEDformer \citep{FEDformer} combines decomposition with frequency-domain learning, and TimeMixer \citep{TimeMixer} exchanges information across temporal scales. Water-quality studies have also coupled wavelet decomposition with long short-term memory (LSTM) to alleviate nonstationarity \citep{chen2024coupled}. However, their fixed temporal assumptions limit adaptation to evolving backgrounds and irregular deviations.

\vspace{-0.2em}
\paragraph{Spatiotemporal Modeling across Monitoring Stations.}
Graph-based models represent monitoring stations as nodes and use geographical, hydrological, or learned relations for cross-station aggregation \citep{graph111,ho2025graph,bi2025long}. STF-GNN \citep{wan2025temporal} combines graph convolution, gate recurrent unit, and self-attention to jointly model spatial and temporal dependencies. Such propagation is useful when stations have meaningful hydrological connectivity. However, in some real-world monitoring networks, stations are sparse, widely separated, and influenced by heterogeneous local environments \citep{zhang2023monitoring,wang2026improving}. Water-quality signals are therefore attenuated, dispersed, or transformed over long travel distances, weakening direct correspondence across monitoring sites. This limits explicit cross-station propagation and motivates station-wise forecasting that exploits spatial information through environmental context.

\vspace{-0.3em}
\paragraph{Rainfall-Informed and Multimodal Water-Quality Forecasting.}

Rainfall is incorporated into water-quality forecasting through process-based and data-driven approaches. Process-based models describe its effects through hydrological processes: SWAT \citep{krysanova2015advances} converts precipitation into runoff and transport fluxes, while WASP \citep{wool2020wasp} simulates pollutant transport and biochemical reactions. Although physically interpretable, they require detailed inputs and site-specific calibration. Data-driven methods learn rainfall effects from observations using lag-selected predictors \citep{ortiz2023methodology} or recurrent networks with meteorological inputs \citep{huang2024riverine}, but commonly reduce rainfall to scalar covariates. Recent multimodal methods preserve spatial information through radar features \citep{baek2020prediction} or direct rainfall-image fusion through CNN \citep{bi2025hybrid}. However, station-specific rainfall events and their duration, relative-lag, and scale-dependent effects remain underexplored.

\section{Methodology}
\label{sec:xfmnet}
We consider multistep and multi-site forecasting with historical in-situ observations and gridded rainfall data. Let
$\mathbf{X}\in\mathbb{R}^{T\times N}$ denote the historical measurements of a target water-quality indicator from $N$ monitoring stations over $T$ time steps, and let
$\mathbf{I}\in\mathbb{R}^{T\times C_I\times H_I\times W_I}$
denote the corresponding gridded rainfall sequence, where $C_I$ is the number of rainfall channels and $H_I\times W_I$ represents the spatial resolution. The directional relationships among monitoring stations are modeled by a directed topology $\mathcal{G}$. Given $\mathbf{X}$, $\mathbf{I}$, and $\mathcal{G}$, our objective is to predict the target indicator at all stations over the next $H$ time steps: $\widehat{\mathbf{Y}}
=
f_{\theta}(\mathbf{X},\mathbf{I},\mathcal{G})
\in\mathbb{R}^{H\times N},$ where $f_{\theta}$ denotes the forecasting model.

\begin{figure*}[htbp]
\centering
\includegraphics[width=0.95\textwidth]{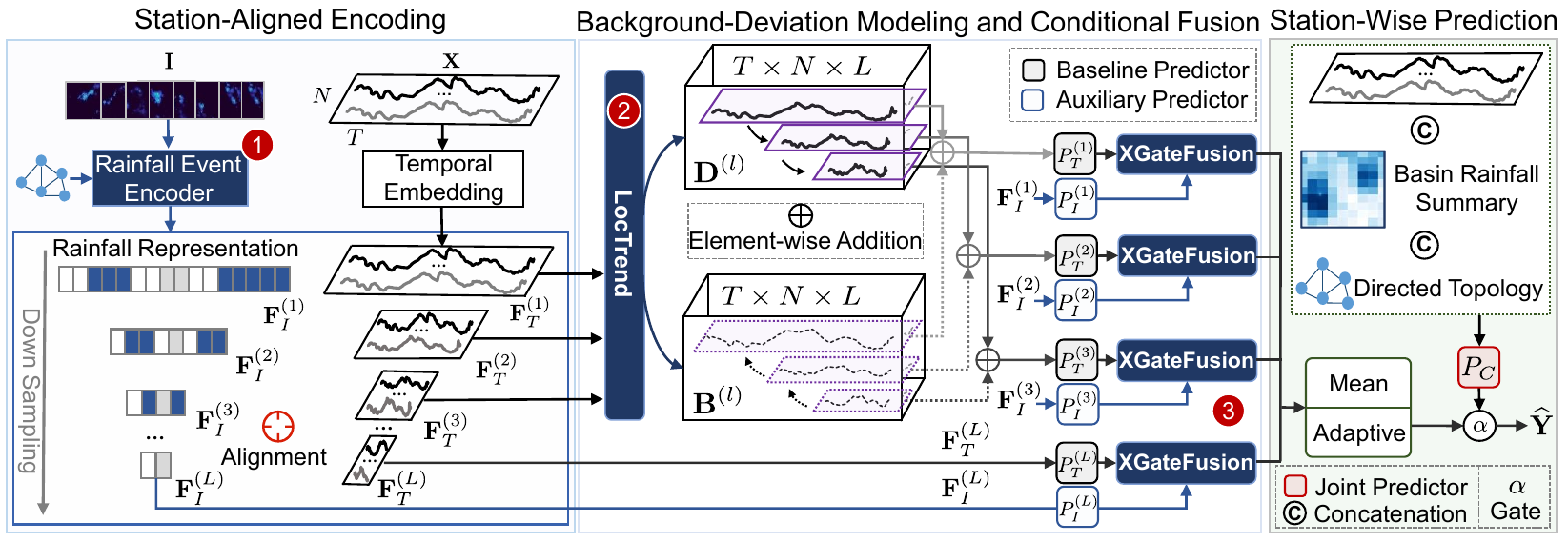}
\caption{\textbf{The RaiNet framework.}
%and they construct aligned multiscale representations, integrate scale-dependent rainfall information, and generate forecasts using multimodal and topology features. 
The three key modules are the rainfall event encoder, \texttt{LocTrend}, and \texttt{XGateFusion}: (1) identifies station-relevant rainfall events; (2) separates evolving backgrounds from dynamic deviations without a fixed period; and (3) conditionally integrates rainfall according to its relative lag and temporal scale.}
\label{fig:framework}
\vspace{-0.6em}
\end{figure*}

% \begin{figure}[!ht]
% \centering
% \includegraphics[width=0.55\columnwidth]{figure/early-fusion.pdf} 
% \caption{\textbf{Early cross-modal interaction via 2D convolution.} Temporal tokens from two modalities are stacked along a ``modality'' axis, and a $(k_t, 2)$ kernel jointly processes local time steps and both modalities to produce fused control tokens.}
% \label{2dconv}
% \end{figure}

%Rather than directly fusing raw rainfall and water-quality observations, RaiNet first constructs modality-specific representations and aligns them at the station level. The overall framework of RaiNet is shown in Figure~\ref{fig:framework}, it extracts rainfall event representations and multiscale water-quality features, which capture complementary dynamics from water-quality evolution and rainfall events. The water-quality representations are further decomposed into structural background and dynamic deviation by \texttt{LocTrend}, enabling the model to distinguish persistent evolution from transient variations. The aligned multimodal features are then integrated through \texttt{XGateFusion}, which captures scale-dependent temporal relationships between the two modalities. Finally, RaiNet integrates the multiscale fused forecasts with a topology-aware prediction branch, which further models inter-station dependencies by incorporating basin-level rainfall summaries and directed station relationships.

The RaiNet architecture is illustrated in Figure~\ref{fig:framework}. It encodes station-oriented rainfall alongside multiscale water-quality dynamics. \texttt{LocTrend} decomposes the water-quality representations into locally evolving structural backgrounds and dynamic deviations. Scale-specific predictors then map both branches into a common forecast space, where \texttt{XGateFusion} captures relative-lag associations and applies rainfall-derived corrections. Finally, the multiscale forecasts are combined with joint prediction that incorporates basin-level rainfall context and directed inter-station relationships.

\subsection{Station-Aligned Encoding}
\label{subsec:embed}
%We construct temporally aligned multiscale representations for water-quality sequences and rainfall grids with $L$ temporal scales.

\paragraph{Multiscale water-quality encoding.}
% Water-quality dynamics are governed by processes evolving at different temporal scales, ranging from short-term fluctuations caused by transient events to long-term variations driven by gradual environmental changes. To capture these heterogeneous temporal patterns, we construct a temporal pyramid through scale-specific downsampling:
% \begin{equation}
% \mathbf{X}^{(l)}=\mathcal{D}_l(\mathbf{X}),
% \qquad
% \mathbf{X}^{(l)}\in\mathbb{R}^{T_l\times N},
% \qquad l\in\{1,\ldots,L\},
% \end{equation}
% where $T_l$ denotes the temporal length at scale $l$. The first scale preserves the original resolution, while subsequent scales are obtained through successive downsampling, providing multiscale representations of water-quality evolution.

Water-quality dynamics evolve at different temporal scales, ranging from short-term fluctuations caused by transient disturbances to long-term variations driven by gradual environmental changes. To capture these heterogeneous patterns, we construct a temporal pyramid through successive average pooling:
\begin{equation}
\mathbf{X}^{(1)}=\mathbf{X},\qquad
\mathbf{X}^{(l)}
=
\operatorname{AvgPool}_{q}\!\left(\mathbf{X}^{(l-1)}\right)
\in\mathbb{R}^{T_l\times N},
\qquad
T_l=\left\lfloor\frac{T_{l-1}}{q}\right\rfloor,
\quad l=2,\ldots,L,
\end{equation}
where $T_l$ is the temporal length at scale $l$, $L$ is the number of scales, and $q$ is the pooling window size. Each scale is then normalized and mapped into a $d$-dimensional latent space as
$\mathbf{F}_T^{(l)}
=\operatorname{Emb}_T(\operatorname{Norm}_l(\mathbf{X}^{(l)}))
\in\mathbb{R}^{T_l\times N\times d}$,
where $\operatorname{Norm}_l(\cdot)$ denotes scale-specific normalization and $\operatorname{Emb}_T(\cdot)$ is a temporal embedding shared across scales. It combines convolutional value embedding with scale-aligned time features, and further details are provided in Appendix~\ref{app:embedding}.
%the station index is omitted for brevity,

\paragraph{Station-oriented rainfall-event encoding.}
%A station-local rainfall value alone cannot distinguish a localized event from part of a basin-wide rainfall system. We interpret local rainfall relative to the simultaneous basin background:
Station locations from $\mathcal{G}$ are first used to align each monitoring station with the rainfall grid. Because station-local rainfall may reflect either a localized event or part of a basin-wide system, we interpret it relative to the simultaneous basin background:
\begin{equation}
r_t^{\mathrm{global}}
=
\frac{\sum_p v_{t,p}I_{t,p}}{\sum_p v_{t,p}},
\qquad
r_{n,t}^{\mathrm{local}}
=
\frac{\sum_p \pi_{n,p}v_{t,p}I_{t,p}}
     {\sum_p \pi_{n,p}v_{t,p}},
\qquad
r_{n,t}^{\mathrm{rel}}
=
r_{n,t}^{\mathrm{local}}-r_t^{\mathrm{global}},
\end{equation}
where $t$, $n$, and $p$ index time steps, monitoring stations, and rainfall-grid cells, respectively; $I_{t,p}$ is the rainfall intensity and $v_{t,p}\in\{0,1\}$ indicates whether it is valid. The normalized spatial prior $\pi_{n,p}$ assigns larger weights to cells closer to station $n$. Accordingly, $r_t^{\mathrm{global}}$, $r_{n,t}^{\mathrm{local}}$, and $r_{n,t}^{\mathrm{rel}}$ denote basin-wide rainfall, station-local rainfall, and their signed spatial contrast. In this case, small contrasts reflect minor spatial variability or measurement uncertainty, and we therefore retain only contrasts exceeding the normal variability observed at each station:
% \begin{equation}
% e_{n,t}
% =
% \operatorname{sign}\!\left(r_{n,t}^{\mathrm{rel}}\right)
% \mathbb{I}\!\left(\left|r_{n,t}^{\mathrm{rel}}\right|>\epsilon_n\right),
% \end{equation}
\begin{equation}
e_{n,t}
=
\operatorname{sign}\!\left(r_{n,t}^{\mathrm{rel}}\right)
\mathbb{I}\!\left(
\left|r_{n,t}^{\mathrm{rel}}\right|>\epsilon_n
\right),
\qquad
e_{n,t}\in\{-1,0,+1\}.
\end{equation}

where $\epsilon_n$ is a station-specific threshold estimated exclusively from the training set, $\operatorname{sign}(\cdot)$ returns the contrast direction, and $\mathbb{I}(\cdot)$ is the indicator function. Thus, $e_{n,t}\in\{-1,0,+1\}$ represents locally weaker rainfall, no significant contrast, or locally stronger rainfall, respectively. This converts raw spatial differences into discrete and directional event states that are comparable across stations.

%A station-relevant event should also persist beyond an isolated hourly change, so the hourly states are aggregated through signed voting within $h$-hour blocks:
A station-relevant event should also persist beyond an isolated observation,
so the time-indexed states are aggregated through signed voting using
persistence setting $h$:
\begin{equation}
\widetilde e_{n,t}
=
\operatorname{sign}
\left(
\sum_{s=b(t)h+1}^{\left(b(t)+1\right)h}e_{n,s}
\right),
\qquad
b(t)=\left\lfloor\frac{t-1}{h}\right\rfloor,
\qquad
\mathbf{F}_I
=
\mathcal{E}_I(\mathbf{I})
=
[\widetilde e_{n,t}]_{t,n},
\end{equation}
%where $b(t)$ identifies the block containing time step $t$, and $\widetilde e_{n,t}$ is the stabilized state assigned to every position in that block. The resulting $\mathbf{F}_I\in\mathbb{R}^{T\times N}$ is the station-related rainfall representation produced by encoder $\mathcal{E}_I$.
where $b(t)$ identifies the block containing time step $t$, and
$\widetilde e_{n,t}$ is assigned to every position in the corresponding
block. The block construction is adapted to each
dataset's sampling interval. The resulting
$\mathbf{F}_I\in\mathbb{R}^{T\times N}$ is the station-related rainfall
representation produced by encoder $\mathcal{E}_I$.

%衔接不好 应该一部分为了对齐水质 一部分有意义
% {\color{red}Furthermore, rainfall events remain informative over different temporal supports, where the dominant event direction is determined by aggregating stabilized states over progressively broader intervals. We apply causal signed-sum pooling using the same aggregation spans as the water-quality pyramid, yielding
% $\mathbf{F}_I^{(l)}=\mathcal{A}_l(\mathbf{F}_I)\in\mathbb{R}^{T_l\times N}$,
% where $\mathcal{A}_l(\cdot)$ sums the event states within each scale-$l$ interval and retains the sign of the result. Consequently, $\mathbf{F}_I^{(l)}$ and $\mathbf{F}_T^{(l)}$ share the same temporal length $T_l$ and form an aligned pair for scale-specific conditional fusion. The overall process is shown in Figure \ref{encoder} and implementation details are provided in Appendix~\ref{app:rain_event}.}

\begin{figure*}[htbp]
\centering
\includegraphics[width=\textwidth]{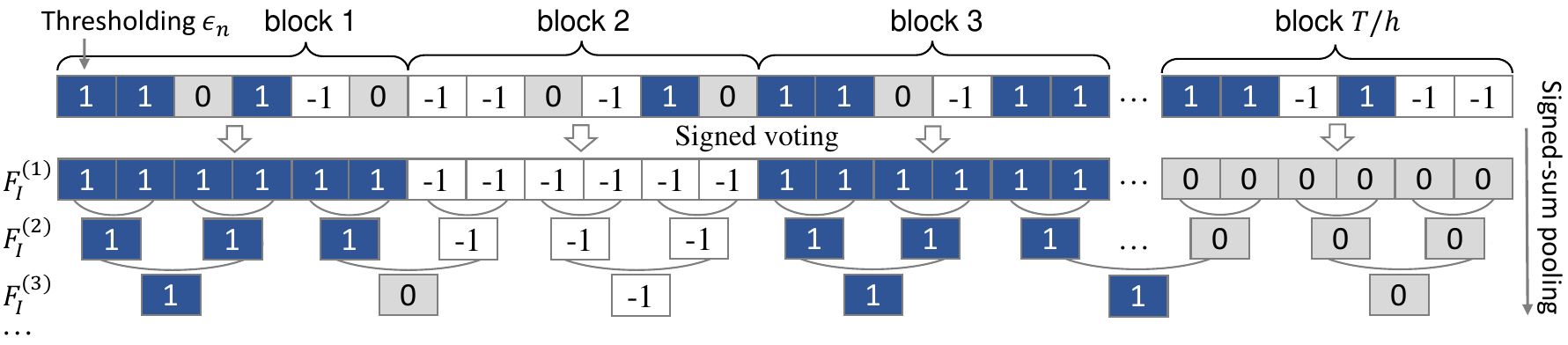}
\caption{\textbf{Rainfall event encoder.} Signed voting with setting $h$ stabilizes each rainfall event, while signed-sum pooling preserves its dominant direction over broader supports and aligns it with the corresponding water-quality scales.} %\textbf{Temporal stabilization and multiscale alignment of rainfall events.}
%\vspace{-1.5em}
\label{encoder}
\end{figure*}

Building on this stabilized event sequence, we further represent rainfall evidence over the temporal supports of the water-quality pyramid. Specifically, causal signed-sum pooling aggregates consecutive states over progressively broader intervals and retains the dominant event direction:
$\mathbf{F}_I^{(l)}
=\mathcal{A}_l(\mathbf{F}_I)
\in\mathbb{R}^{T_l\times N}$,
where $\mathcal{A}_l(\cdot)$ uses the aggregation span of scale $l$. This operation both captures the persistence of rainfall evidence at different temporal scales and aligns $\mathbf{F}_I^{(l)}$ with $\mathbf{F}_T^{(l)}$ along the station and temporal dimensions. The overall process is illustrated in Figure~\ref{encoder}, with implementation details provided in Appendix~\ref{app:rain_event}.

\subsection{Background-Deviation Modeling and Conditional Fusion}
\label{subsec:decomp_fusion}
The two modalities play asymmetric roles in RaiNet. $\mathbf{F}_T^{(l)}$ captures intrinsic water-quality dynamics, whereas $\mathbf{F}_I^{(l)}$ provides temporally and station-aligned rainfall-event conditions. This module first decomposes water-quality dynamics into a locally evolving background and dynamic deviations, and then selectively applies lag-aware rainfall corrections to the resulting baseline.

\subsubsection{LocTrend Background-Deviation Decomposition}
Fixed-period decomposition imposes recurrent temporal structures that do not match the locally evolving patterns of water-quality dynamics. \texttt{LocTrend} reconstructs a locally evolving structural background from data-dependent patterns without specifying a daily, weekly, or annual period.

\begin{wrapfigure}{r}{0.43\columnwidth}
    \vspace{-2em}
    \centering
    \includegraphics[width=\linewidth]{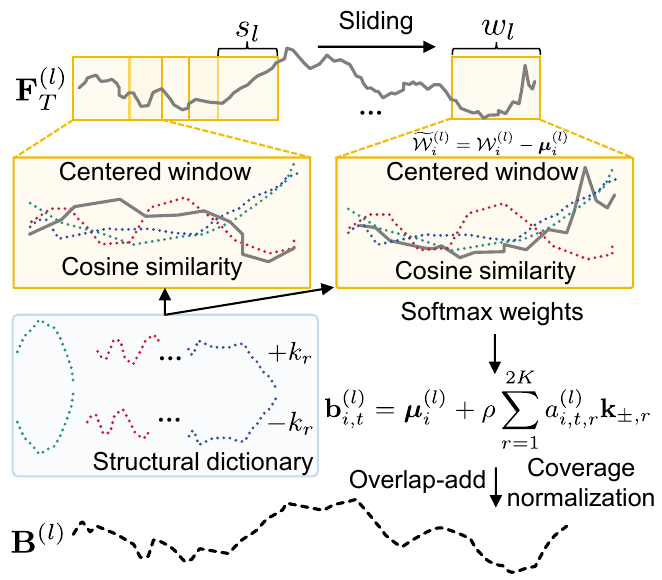}
    \caption{\texttt{LocTrend} reconstructs local structural backgrounds.} %from overlapping windows using a training-fitted dictionary
    \label{fig:loctrend}
    \vspace{-1.1em}
\end{wrapfigure}

\texttt{LocTrend} is applied only before the temporal sequence becomes excessively compressed. Let $L_{\mathrm{dir}}$ denote the number of coarsest scales preserved through the direct temporal path. It is applied to the remaining scales ($l=1,\ldots,L-L_{\mathrm{dir}}$):
\begin{equation}
\big(\mathbf{B}^{(l)},\mathbf{D}^{(l)}\big)
=
\operatorname{LocTrend}\!\left(\mathbf{F}_T^{(l)}\right).
\label{eq:loctrend_scale_routing}
\end{equation}
The remaining coarsest scales
$l=L-L_{\mathrm{dir}}+1,\ldots,L$
are preserved without further decomposition, as their short sequences already summarize broad temporal contexts and may lose useful coarse-scale information under local decomposition. 

For each decomposed scale, we extract overlapping windows and remove their local levels:
\begin{equation}
\mathcal{W}_i^{(l)}
=
\mathbf{F}_{T,\,q_i:q_i+w_l-1}^{(l)},
\qquad
\boldsymbol{\mu}_i^{(l)}
=
\frac{1}{w_l}\sum_{t=1}^{w_l}\mathcal{W}_{i,t}^{(l)},
\qquad
\widetilde{\mathcal{W}}_i^{(l)}
=
\mathcal{W}_i^{(l)}-\boldsymbol{\mu}_i^{(l)},
\label{eq:loc_window}
\end{equation}
where $w_l$ and $s_l$ are the window length and stride, respectively, and $q_i=(i-1)s_l+1$ is the starting position of window $i$. Centering shifts the dictionary from absolute levels to local patterns.

\paragraph{Training-fitted structural dictionary.} We summarize recurrent structures observed in the training windows as a shared dictionary
$\mathbf{K}=\{\mathbf{k}_r\}_{r=1}^{K}$.
It is fitted once using a two-pass structural initialization:
\begin{equation}
\left\{
\widetilde{\mathcal{W}}_q
\right\}_{q\in\mathcal{Q}_{\mathrm{tr}}}
\xrightarrow[\text{averaged kernel eigendecomposition}]
{\text{RBF temporal similarities}}
\left\{(\lambda_r,\mathbf{u}_r)\right\}_{r=1}^{K}
\xrightarrow[\text{feature covariance eigendecomposition}]
{\text{window projection}}
\mathbf{K},
\label{eq:loctrend_dictionary}
\end{equation}
where $\mathcal{Q}_{\mathrm{tr}}$ is the collection of centered training windows, and $K$ is the number of retained components and resulting dictionary atoms. In the first pass, a radial-basis-function (RBF) kernel converts pairwise distances within each window into nonlinear temporal similarities; averaging their centered Gram matrices and retaining the top eigenpairs
$\{(\lambda_r,\mathbf{u}_r)\}_{r=1}^{K}$
identifies temporal directions recurring across training windows. In the second pass, the windows are projected along these directions, and the leading eigenvectors of their aggregated feature covariance form the atoms $\mathbf{k}_r$. The resulting dictionary is shared across the decomposed scales and fixed during ordinary training. Appendix~\ref{app:loctrend_init} details the adaptive RBF similarity construction, temporal eigendirection extraction, and their conversion into feature-space structural atoms.

\paragraph{Local background reconstruction.}
To represent centered patterns in either direction, we form the symmetric dictionary
$\mathbf{K}_{\pm}=[\mathbf{K};-\mathbf{K}]$ with $2K$ atoms. Each centered token is softly assigned using temperature-scaled cosine similarity:
\begin{equation}
a_{i,t,r}^{(l)}
=
\operatorname{softmax}_{r}
\left(
\frac{
\operatorname{cos}\!\left(
\widetilde{\mathbf{w}}_{i,t}^{(l)},
\mathbf{k}_{\pm,r}
\right)
}{\vartheta}
\right),
\qquad
\mathbf{b}_{i,t}^{(l)}
=
\boldsymbol{\mu}_i^{(l)}
+
\rho\sum_{r=1}^{2K}
a_{i,t,r}^{(l)}\mathbf{k}_{\pm,r},
\label{eq:loc_reconstruction}
\end{equation}
where $\widetilde{\mathbf{w}}_{i,t}^{(l)}$ is token $t$ of the centered window, $\vartheta$ is the assignment temperature, and $\rho$ controls the centered reconstruction strength. This soft assignment allows the background to follow locally recurring shapes without imposing a predefined period. Finally, overlapping reconstructions are averaged according to their coverage to obtain $\mathbf{B}^{(l)}$ and $\mathbf{D}^{(l)} = \mathbf{F}_T^{(l)}-\mathbf{B}^{(l)}$. Thus, $\mathbf{B}^{(l)}$ captures the locally evolving structural background, while $\mathbf{D}^{(l)}$ preserves the exact dynamic remainder. The complete LocTrend forward procedure is detailed in Appendix~\ref{locccccc}. By centering local windows and restoring their means, \texttt{LocTrend} absorbs local level shifts into the background while preserving dynamic deviations, without relying on fixed periodic structures (Appendix~\ref{app:loctrend_theory}).

\subsubsection{Cross-Scale Modeling and Scale-Specific Prediction}
To improve consistency across decomposed scales, RaiNet propagates dynamic deviations from fine to coarse resolutions (arrows between different scales in Figure~\ref{fig:framework}), where short-term changes are more clearly resolved, and structural backgrounds from coarse to fine resolutions, where long-range context is more stable, i.e.,
$\{\overline{\mathbf{D}}^{(l)}\}
=\mathcal{M}_{D}^{\mathrm{f2c}}(\{\mathbf{D}^{(l)}\})$
and
$\{\overline{\mathbf{B}}^{(l)}\}
=\mathcal{M}_{B}^{\mathrm{c2f}}(\{\mathbf{B}^{(l)}\})$. Each transition adaptively combines convolutional and temporal linear paths. Details are provided in Appendix~\ref{app:cross_scale_mixing}. 

For each decomposed scale, each pair is recombined as
$\widetilde{\mathbf{F}}_{T}^{(l)}
=\mathbf{F}_{T}^{(l)}
+\operatorname{FFN}(\overline{\mathbf{B}}^{(l)}
+\overline{\mathbf{D}}^{(l)})$.
The remaining coarsest scales bypass this operation.
%such that $\widetilde{\mathbf{F}}_{T}^{(l)}=\mathbf{F}_{T}^{(l)}$ for $l>L-L_{\mathrm{dir}}$.
Scale-specific predictors then map the two modalities into the common water-quality prediction space:
\begin{equation}
\widehat{\mathbf{Y}}_{T}^{(l)}
=
P_T^{(l)}\!\left(\widetilde{\mathbf{F}}_{T}^{(l)}\right),
\qquad
\widehat{\mathbf{Y}}_{I}^{(l)}
=
P_I^{(l)}\!\left(\mathbf{F}_{I}^{(l)}\right),
\qquad
\widehat{\mathbf{Y}}_{T}^{(l)},
\widehat{\mathbf{Y}}_{I}^{(l)}
\in\mathbb{R}^{H\times N}.
\label{eq:scale_specific_predictors}
\end{equation}
where $\widehat{\mathbf{Y}}_{T}^{(l)}$ is the predicted water-quality baseline. $P_I^{(l)}$ translates historical rainfall events into a candidate water-quality correction in the forecast space, yielding $\widehat{\mathbf{Y}}_{I}^{(l)}$. \texttt{XGateFusion} then determines how much of this correction should be applied for each station, forecast step, and scale.

\subsubsection{XGateFusion}
Rainfall correction should only be introduced when compatible with the current water-quality state.
%\vspace{-1em}
\begin{wrapfigure}{l}{0.4\columnwidth}
\centering
\includegraphics[width=\linewidth]{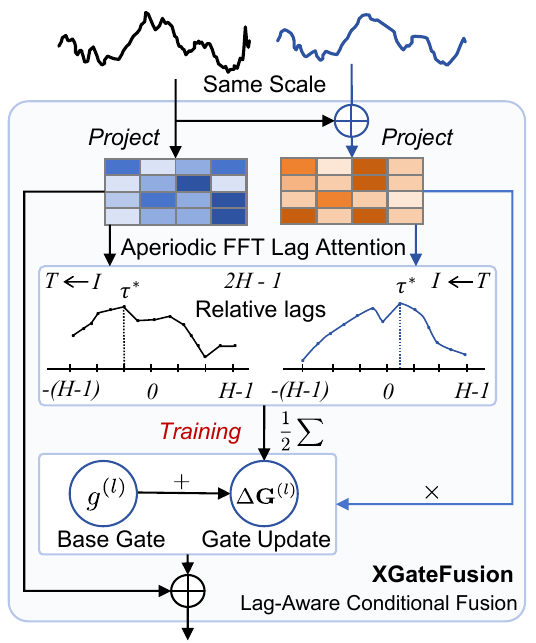}
\caption{Bidirectional aperiodic FFT lag attention adapts corrections across scales, stations, and forecast steps.} %across scales, stations, and forecast steps
\label{fig:xgate}
\vspace{-1.8em}
\end{wrapfigure}
At scale $l$, \texttt{XGateFusion} compares the station-wise baseline
$\mathbf{u}_{T,n}^{(l)}=\widehat{\mathbf{y}}_{T,n}^{(l)}$
with its rainfall-conditioned counterpart
$\mathbf{u}_{I,n}^{(l)}
=\widehat{\mathbf{y}}_{T,n}^{(l)}
+\widehat{\mathbf{y}}_{I,n}^{(l)}$,
where $\widehat{\mathbf{y}}_{T,n}^{(l)}$ and
$\widehat{\mathbf{y}}_{I,n}^{(l)}$ are the station-$n$ columns of the scale-level baseline and correction, respectively. Comparing them in the same prediction space allows the model to evaluate candidate corrections without directly matching heterogeneous representations.

Rainfall-related changes may align with the baseline at different relative forecast offsets. 
Bidirectional aperiodic lag attention therefore evaluates their statistical associations over the complete lag range, while avoiding artificial wraparound at the horizon boundaries. The interactions are computed by Fast Fourier Transform (FFT) with $\mathcal{O}(H\log H)$ temporal complexity, without explicitly constructing the full lag-interaction matrix. The resulting representation produces a bounded gate update
$\Delta\mathbf{G}^{(l)}\in\mathbb{R}^{H\times N}$,
allowing the correction strength to vary across stations and forecast steps. Together with the scale-level base gate $g^{(l)}$, the fused prediction is
\begin{equation}
\widehat{\mathbf{Y}}_{TI}^{(l)}
=
\widehat{\mathbf{Y}}_{T}^{(l)}
+
\left(
g^{(l)}\mathbf{1}_{H\times N}
+
\Delta\mathbf{G}^{(l)}
\right)
\odot
\widehat{\mathbf{Y}}_{I}^{(l)}.
\label{eq:xgate_output}
\end{equation}
where $\mathbf{1}_{H\times N}$ is an all-one matrix and $\odot$ denotes element-wise multiplication. The base gate $g^{(l)}$ controls the scale-level contribution of rainfall, whereas $\Delta\mathbf{G}^{(l)}$ provides station and horizon-specific adaptation. The dynamic gate-update head is initialized to zero for stable training. Detailed derivations of \texttt{XGateFusion} are provided in Appendix~\ref{app:xgate}.

\subsection{Station-Wise Prediction}
\label{subsec:station_prediction}
Because the informative temporal scale varies across samples, RaiNet uses forecast statistics to adaptively weight the scale-wise predictions, yielding
$\widehat{\mathbf{Y}}_{\mathrm{w}}
=\sum_{l=1}^{L}\omega_l\widehat{\mathbf{Y}}_{TI}^{(l)}$.
It is balanced with their mean
$\widehat{\mathbf{Y}}_{\mathrm{m}}
=L^{-1}\sum_{l=1}^{L}\widehat{\mathbf{Y}}_{TI}^{(l)}$, i.e., $\widehat{\mathbf{Y}}_{TI}
=
\zeta_{\mathrm{m}}\widehat{\mathbf{Y}}_{\mathrm{m}}
+
\zeta_{\mathrm{w}}\widehat{\mathbf{Y}}_{\mathrm{w}},$
and $\zeta_{\mathrm{m}}+\zeta_{\mathrm{w}}=1,$ where the sample-dependent weights balance stable averaging against selective scale aggregation.

Beyond station-aligned rainfall corrections, water-quality prediction also depends on basin-level rainfall and spatial relations among stations. The Joint Predictor $P_C$ concatenates the encoded water-quality history, basin rainfall summaries, and topological features, allowing their complementary information to be modeled jointly. It produces
$\widehat{\mathbf{Y}}_C
=P_C(\mathbf{X},\mathbf{c}_I,\mathcal{G})
\in\mathbb{R}^{H\times N}$,
where $\mathbf{c}_I$ denote the basin rainfall context. The final prediction is $\widehat{\mathbf{Y}}
=
(1-\alpha)\widehat{\mathbf{Y}}_{TI}
+
\alpha\widehat{\mathbf{Y}}_C,
\alpha\in[0,1].$ %This retains the multiscale prediction as an anchor while incorporating complementary basin-level and topological context.

\section{Experiments}
\label{sec:experiments}

\subsection{Experimental Setup}
% We evaluate RaiNet on three hydrological basins, where stations with sufficient temporal coverage form basin-wide monitoring networks. The datasets contain five-year hourly turbidity records from 15 stations, three-year hourly nitrate records from five stations, and five-year 30-minute nitrate records from five stations, respectively, all aligned with gridded precipitation. We publicly release the resulting benchmark of over 150,000 paired time steps. Each dataset is chronologically split into training, validation, and test sets at a 7:1:2 ratio. Further details are provided in Appendix~\ref{sec:appendix_dataset}.
We evaluate RaiNet on three basin-level multimodal datasets comprising hydrologically connected monitoring stations. They contain five-year hourly turbidity records from 15 stations, three-year hourly nitrate records from five stations, and five-year 30-minute nitrate records from five stations, all aligned with gridded precipitation. We publicly release over 150,000 paired time steps as benchmarks. Each dataset is chronologically split at 7:1:2, with details in Appendix~\ref{sec:appendix_dataset}.

RaiNet uses \(L=4\) temporal scales, constructed by successive average pooling
with window size \(q=2\). We set \(L_{\mathrm{dir}}=1\), so the last coarsest scale follows the direct
temporal path. Rainfall-event states %are %stabilized using six-hour persistence blocks before multiscale alignment. %$h=6$ represents short sub-daily event persistence: this interval is long enough to capture sustained spatial rainfall organization potentially associated with initial runoff responses, while remaining short enough to avoid merging temporally separate events. 
$h=6$ captures short sub-daily rainfall persistence without merging separate events.
For \texttt{LocTrend}, we use
\(w_l=24\), \(s_l=12\), \(K=4\), \(\vartheta=1\), and \(\rho=1\). We set
the hidden dimension \(d=16\), the feedforward dimension to 32, and dropout
to 0.1. Models are trained with a batch size of 64 and a learning
rate of \(10^{-3}\). Sensitivity analysis of hyperparameters is
provided in Appendix~\ref{sen}.

We compare RaiNet with four groups of baselines. Time-series models include TimeKAN \citep{TimeKAN}, FilterTS \citep{FilterTS}, GTR \citep{cao2026gtr}, and MixLinear \citep{ma2026mixlinear}. Spatio-temporal models include HyperD \citep{shao2026hyperd}, RSTIB-MLP \citep{chen2025information}, and ST-SSDL \citep{gao2026different}, representing periodicity decoupling, robust spatio-temporal learning, and deviation-aware self-supervision. Water-quality-specific baselines include PIWON \citep{tang2026non}, D-TGCN \citep{zhang2026incorporating}, CMLIP \citep{bi2025hybrid}, RGCN \citep{smith2024predictive}, and XGBR \citep{ali2025near}. We also compare with diffusion-based SimDiff \citep{ding2026simdiff} and MA-TSD \citep{wang2025non}. Details are provided in Appendix~\ref{cm}.

\subsection{Main Results}\label{sec:res}

As shown in Tables~\ref{tab:water_quality_long_horizon} and
\ref{tab:water_quality_long_horizon2}, RaiNet consistently achieves the best
MSE across all datasets and horizons, reducing it by up to 21.6\% over the
strongest baseline. It also obtains the best MAE in eight of nine settings;
the only exception is 170900 at 192 steps, where RSTIB-MLP is marginally
better. The gains remain consistent across different indicators, sampling
intervals, and basin configurations, demonstrating consistent robustness.
In contrast, conventional spatiotemporal models are unstable, particularly
on 071200 and 051201. Water-quality stations are sparsely distributed and
exhibit heterogeneous, asynchronous dynamics, making generic cross-station
propagation prone to introducing irrelevant spatial information. RaiNet
avoids this mismatch by forecasting station-wise dynamics while using spatial
rainfall patterns as conditional environmental context, exploiting
basin information without enforcing direct propagation between stations. It
also consistently outperforms water-quality-specific and diffusion-based
models. Full-horizon results, detailed analyses, and computational costs are
provided in Appendix~\ref{res}.

\begin{table*}[htbp]
\centering
\caption{
Forecasting results with representative time-series and spatiotemporal baselines.
The best and second-best results are highlighted in \textbf{bold} and
\underline{underline}, respectively.
}
\label{tab:water_quality_long_horizon}
\small
\setlength{\tabcolsep}{1.8pt}
\renewcommand{\arraystretch}{1.2}
\begin{adjustbox}{max width=\textwidth}
\begin{tabular}{
@{}cc|cc|cc|cc|cc|cc|cc|cc|cc
@{}}
\toprule
\multicolumn{2}{c|}{\textbf{Model}}
&
\multicolumn{2}{c|}{\textbf{TimeKAN}}
&
\multicolumn{2}{c|}{\textbf{FilterTS}}
&
\multicolumn{2}{c|}{\textbf{GTR}}
&
\multicolumn{2}{c|}{\textbf{MixLinear}}
&
\multicolumn{2}{c|}{\textbf{HyperD}}
&
\multicolumn{2}{c|}{\textbf{RSTIB-MLP}}
&
\multicolumn{2}{c|}{\textbf{ST-SSDL}}
&
\multicolumn{2}{c}{\textbf{RaiNet}}
\\
\cmidrule(lr){3-4}
\cmidrule(lr){5-6}
\cmidrule(lr){7-8}
\cmidrule(lr){9-10}
\cmidrule(lr){11-12}
\cmidrule(lr){13-14}
\cmidrule(lr){15-16}
\cmidrule(lr){17-18}
\multicolumn{2}{c|}{\textbf{Metric}}
&
\textbf{MSE} & \textbf{MAE}
&
\textbf{MSE} & \textbf{MAE}
&
\textbf{MSE} & \textbf{MAE}
&
\textbf{MSE} & \textbf{MAE}
&
\textbf{MSE} & \textbf{MAE}
&
\textbf{MSE} & \textbf{MAE}
&
\textbf{MSE} & \textbf{MAE}
&
\textbf{MSE} & \textbf{MAE}
\\
\midrule

\multirow{3}{*}{
\rotatebox[origin=c]{90}{\textbf{\texttt{170900}}}
}
& 24
& \underline{0.2628} & \underline{0.1667}
& 0.2692 & 0.1722
& 0.2763 & 0.1726
& 0.3790 & 0.2324
& 0.3786 & 0.1909
& 0.2960 & 0.1871
& 0.2792 & 0.2107
& \textbf{0.2185} & \textbf{0.1560}
\\

& 96
& 0.4778 & \underline{0.2623}
& \underline{0.4771} & 0.2626
& 0.4834 & 0.2648
& 0.5739 & 0.3208
& 0.5518 & 0.2729
& 0.4994 & 0.2676
& 0.5410 & 0.3177
& \textbf{0.4096} & \textbf{0.2604}
\\

& 192
& 0.6053 & 0.3204
& 0.5974 & 0.3199
& 0.5975 & 0.3207
& 0.7047 & 0.3786
& 0.6442 & 0.3183
& \underline{0.5937} & \textbf{0.3061}
& 0.6603 & 0.3691
& \textbf{0.5283} & \underline{0.3102}
\\

\midrule

\multirow{3}{*}{
\rotatebox[origin=c]{90}{\textbf{\texttt{071200}}}
}
& 24
& \underline{0.1084} & 0.1905
& 0.1105 & \underline{0.1885}
& 0.1162 & 0.1967
& 0.2458 & 0.2848
& 0.6530 & 0.4605
& 0.1630 & 0.2555
& 0.1516 & 0.2535
& \textbf{0.0872} & \textbf{0.1747}
\\

& 96
& \underline{0.4412} & 0.3802
& 0.4473 & \underline{0.3784}
& 0.4591 & 0.3894
& 0.5818 & 0.4424
& 0.7663 & 0.5291
& 0.4613 & 0.3912
& 0.7145 & 0.5240
& \textbf{0.3225} & \textbf{0.3344}
\\

& 192
& 0.6771 & \underline{0.4918}
& 0.6879 & 0.4944
& 0.7010 & 0.5041
& 0.8131 & 0.5463
& 0.8453 & 0.5759
& \underline{0.6597} & 0.4930
& 1.0558 & 0.7019
& \textbf{0.5415} & \textbf{0.4443}
\\

\midrule

\multirow{3}{*}{
\rotatebox[origin=c]{90}{\textbf{\texttt{051201}}}
}
& 24
& \underline{0.0074} & \underline{0.0328}
& 0.0075 & 0.0337
& 0.0081 & 0.0353
& 0.0194 & 0.0599
& 1.3482 & 0.7703
& 0.7583 & 0.6066
& 0.0217 & 0.0821
& \textbf{0.0072} & \textbf{0.0318}
\\

& 96
& \underline{0.0462} & \underline{0.0889}
& 0.0468 & 0.0903
& 0.0491 & 0.0938
& 0.0609 & 0.1089
& 1.3572 & 0.8016
& 0.7660 & 0.6449
& 0.0756 & 0.1626
& \textbf{0.0414} & \textbf{0.0867}
\\

& 192
& 0.1129 & 0.1524
& \underline{0.1113} & \underline{0.1505}
& 0.1151 & 0.1536
& 0.1378 & 0.1763
& 1.3842 & 0.8280
& 0.9683 & 0.7014
& 0.1491 & 0.2120
& \textbf{0.0992} & \textbf{0.1455}
\\

\bottomrule
\end{tabular}
\end{adjustbox}
\end{table*}

\subsection{Ablation Studies}
\begin{wrapfigure}{r}{0.65\columnwidth}
    \vspace{-1.2em}
    \centering
    \includegraphics[
        width=\linewidth
    ]{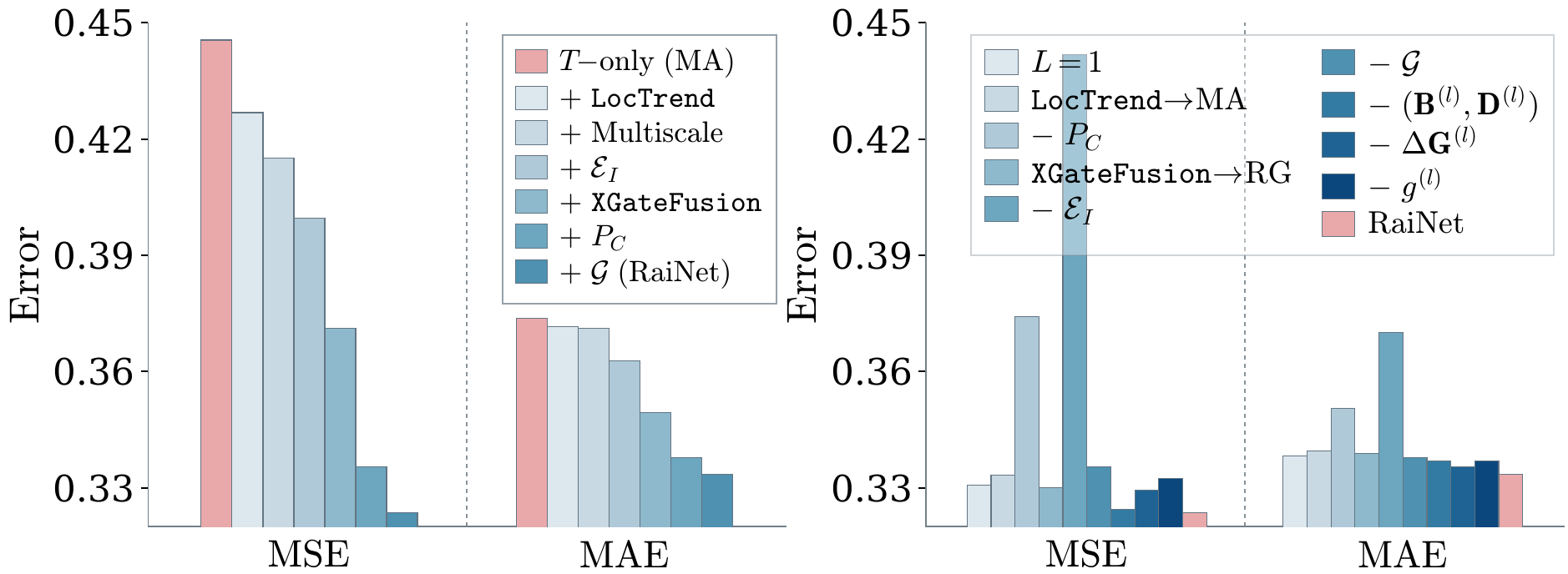}
    \caption{Cumulative and standalone ablations of RaiNet.}
    \label{aba}
    %\vspace{-1.4em}
\end{wrapfigure}
Figure~\ref{aba} evaluates RaiNet through cumulative additions (left) and standalone variants (right), using 071200 as a representative setting. MA and RG denote moving average and residual gate, respectively; $L=1$ removes multiscale modeling, while ``$-$'' and ``$\rightarrow$'' denote removal and replacement. Starting from the temporal-only baseline, progressively introducing \texttt{LocTrend}, multiscale modeling, $\mathcal{E}_I$, \texttt{XGateFusion}, $P_C$, and $\mathcal{G}$ consistently reduces both errors, lowering MSE from 0.446 to 0.324 and MAE from 0.374 to 0.333. This monotonic improvement shows that the components contribute progressively instead of relying on a single dominant modification. The standalone variants provide complementary evidence: every removal or replacement increases both metrics, with particularly clear degradation after removing $\mathcal{E}_I$ or $P_C$. The replacement tests further demonstrate the effectiveness of adaptive decomposition and conditional gating. These results verify both the individual effectiveness and the cumulative compatibility of the proposed components. Further analyses are provided in Appendix~\ref{abala}.

\subsection{Qualitative Analysis}
\label{QA}
% \vspace{-1.7em}
% \begin{wrapfigure}{l}{0.4\columnwidth}
% \centering
% \includegraphics[width=\linewidth]{figure/locloc.pdf}
% \caption{\texttt{LocTrend} decomposition analysis.} %\texttt{LocTrend} produces decompositions by assigning different weights and signs to a shared dictionary.
% \label{locloc}
% \vspace{-2.5em}
% \end{wrapfigure}

\begin{wrapfigure}{r}{0.4\columnwidth}
    \vspace{-2.3em}
    \centering
    \includegraphics[
        width=\linewidth
    ]{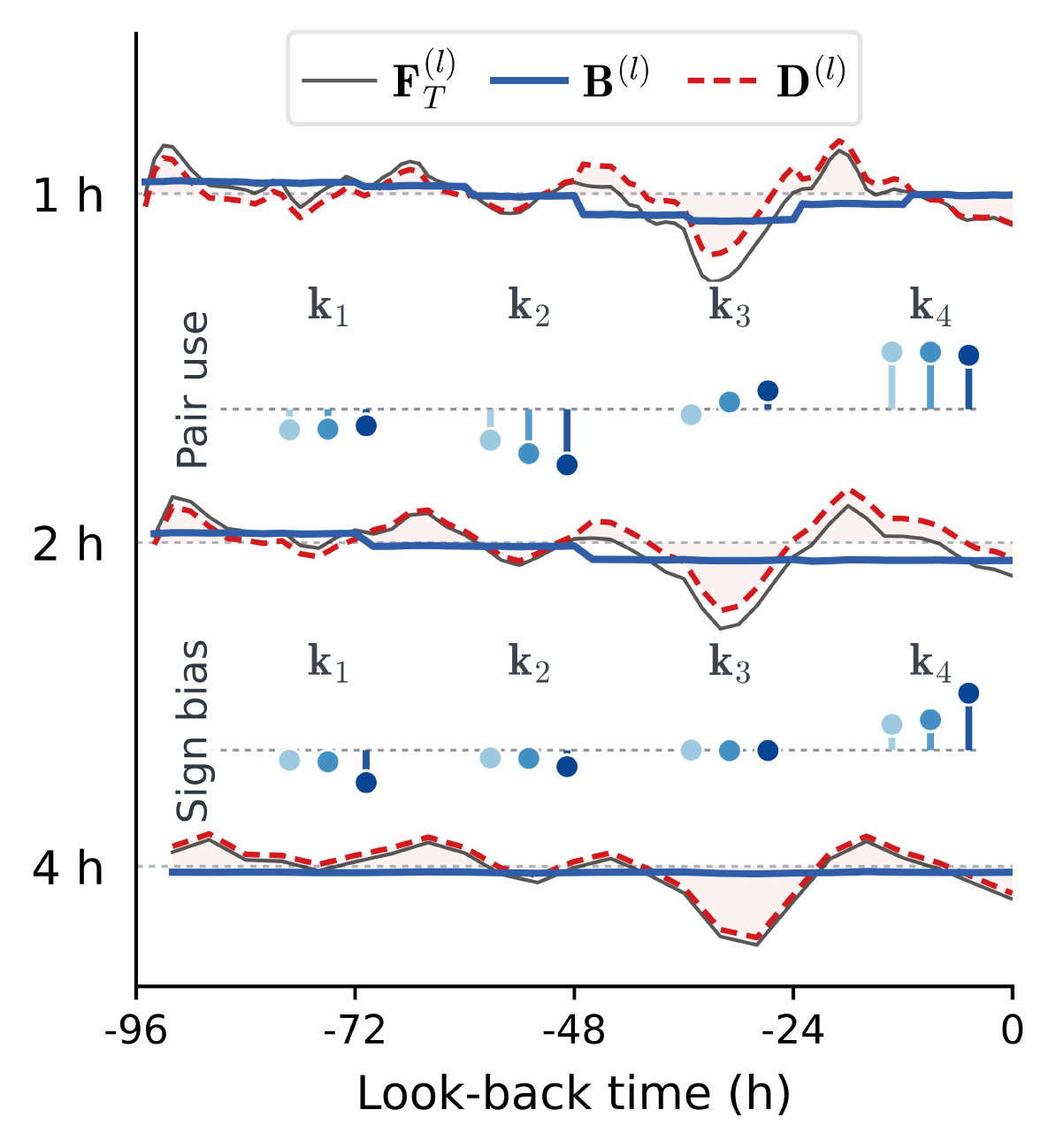}
    \caption{\texttt{LocTrend} analysis.}
    \label{locloc}
    \vspace{-1.9em}
\end{wrapfigure}

Figure~\ref{locloc} analyzes \texttt{LocTrend} at the 1, 2, and 4-hour scales. The line plots show that the same temporal representation is decomposed into different locally evolving backgrounds and exact dynamic deviations as the resolution changes: finer scales retain more local structural variation, whereas the background becomes progressively smoother at coarser scales. The lollipop plots further explain this adaptation. \emph{Pair use} measures the relative utilization of each positive-negative dictionary pair, while \emph{Sign bias} indicates the preference between its two signs. Although all scales share the same structural dictionary, their distinct assignment patterns show that \texttt{LocTrend} selectively recombines shared structural atoms according to the temporal resolution. This shared-dictionary yet scale-adaptive design enables flexible background-deviation decomposition without introducing scale-specific dictionaries or predefined temporal periods. This enables irregular but recurrent local structures to be absorbed into the background while preserving transient departures in the deviation branch. The nearly flat background at the coarsest scale further supports bypassing decomposition after additional downsampling.

\begin{figure}[htbp]
    \centering
    \includegraphics[width=\linewidth]{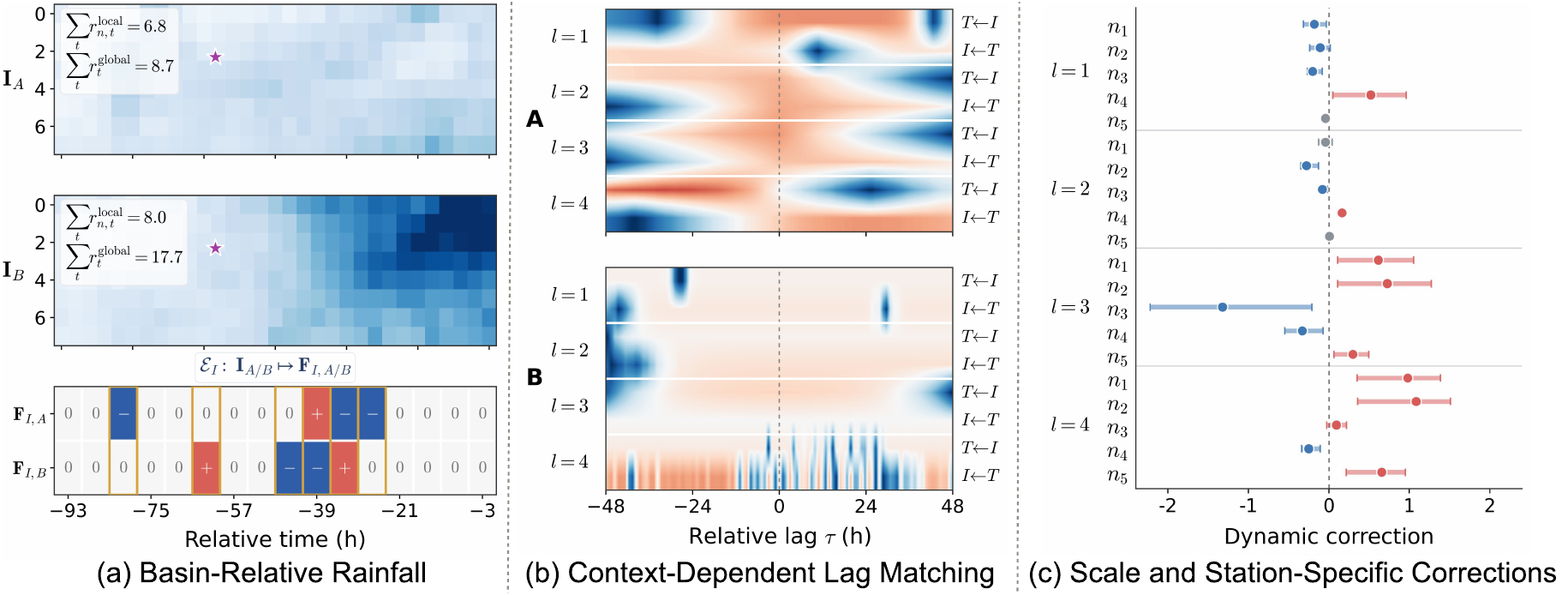}
\caption{\textbf{Analysis of the rainfall event encoder and \texttt{XGateFusion}.}
(a) $\mathcal{E}_I$ distinguishes station-aligned events under different basin contexts.
(b) \texttt{XGateFusion} learns context and scale-dependent relative-lag profiles.
(c) The resulting corrections to the baseline predictions vary across scales and stations; points and intervals denote their means and 10-90th percentiles.}
    \label{fig:rain_xgate_analysis}
\end{figure}

Figure~\ref{fig:rain_xgate_analysis} shows how RaiNet progressively transforms rainfall into conditional forecast corrections. The rainfall event encoder distinguishes similar station-local rainfall under different basin-wide contexts and produces different station-aligned event states instead of relying on absolute rainfall intensity alone (a). Given the same rainfall-event representation but different water-quality histories, \texttt{XGateFusion} learns distinct bidirectional relative-lag profiles across temporal scales, demonstrating context-dependent statistical alignment (b). These profiles are subsequently translated into signed corrections whose direction and magnitude vary across scales and stations (c). Additional qualitative analyses are provided in Appendix~\ref{qa}.%Overall, RaiNet interprets rainfall relative to both basin and water-quality contexts and selectively adjusts the baseline forecast.

\begin{table*}[t]
\centering
\caption{
Forecasting results with water-quality-specific and diffusion-based models.
The best and second-best results are highlighted in \textbf{bold} and
\underline{underline}, respectively.
}
\label{tab:water_quality_long_horizon2}
\small
\setlength{\tabcolsep}{1.8pt}
\renewcommand{\arraystretch}{1.2}
\begin{adjustbox}{max width=\textwidth}
\begin{tabular}{
@{}cc|cc|cc|cc|cc|cc|cc|cc|cc
@{}}
\toprule
\multicolumn{2}{c|}{\textbf{Model}}
&
\multicolumn{2}{c|}{\textbf{PIWON}}
&
\multicolumn{2}{c|}{\textbf{D-TGCN}}
&
\multicolumn{2}{c|}{\textbf{CMLIP}}
&
\multicolumn{2}{c|}{\textbf{RGCN}}
&
\multicolumn{2}{c|}{\textbf{XGBR}}
&
\multicolumn{2}{c|}{\textbf{SimDiff}}
&
\multicolumn{2}{c|}{\textbf{MA-TSD}}
&
\multicolumn{2}{c}{\textbf{RaiNet}}
\\
\cmidrule(lr){3-4}
\cmidrule(lr){5-6}
\cmidrule(lr){7-8}
\cmidrule(lr){9-10}
\cmidrule(lr){11-12}
\cmidrule(lr){13-14}
\cmidrule(lr){15-16}
\cmidrule(lr){17-18}
\multicolumn{2}{c|}{\textbf{Metric}}
&
\textbf{MSE} & \textbf{MAE}
&
\textbf{MSE} & \textbf{MAE}
&
\textbf{MSE} & \textbf{MAE}
&
\textbf{MSE} & \textbf{MAE}
&
\textbf{MSE} & \textbf{MAE}
&
\textbf{MSE} & \textbf{MAE}
&
\textbf{MSE} & \textbf{MAE}
&
\textbf{MSE} & \textbf{MAE}
\\
\midrule

\multirow{3}{*}{
\rotatebox[origin=c]{90}{\textbf{\texttt{170900}}}
}
& 24
& \underline{0.2523} & 0.2013
& 0.2719 & 0.2111
& 0.2796 & 0.1794
& 0.2976 & 0.2113
& 0.3924 & 0.1913
& 0.2773 & \underline{0.1641}
& 0.2843 & 0.2023
& \textbf{0.2185} & \textbf{0.1560}
\\

& 96
& 0.5303 & 0.3326
& \underline{0.4633} & 0.3072
& 0.4943 & 0.2723
& 0.4795 & 0.3179
& 0.8964 & 0.3228
& 0.5062 & \underline{0.2643}
& 0.5011 & 0.3314
& \textbf{0.4096} & \textbf{0.2604}
\\

& 192
& 0.9224 & 0.5161
& \underline{0.5618} & 0.3661
& 0.6139 & 0.3338
& 0.5669 & 0.3673
& 1.4016 & 0.4131
& 0.6387 & \underline{0.3250}
& 0.6263 & 0.4098
& \textbf{0.5283} & \textbf{0.3102}
\\

\midrule

\multirow{3}{*}{
\rotatebox[origin=c]{90}{\textbf{\texttt{071200}}}
}
& 24
& \underline{0.0961} & \underline{0.1936}
& 0.1166 & 0.2043
& 0.1272 & 0.2011
& 0.1784 & 0.2460
& 0.1227 & 0.1999
& 0.1387 & 0.2068
& 0.1280 & 0.2168
& \textbf{0.0872} & \textbf{0.1747}
\\

& 96
& \underline{0.4115} & \underline{0.3869}
& 0.4347 & 0.4086
& 0.4734 & 0.3919
& 0.4476 & 0.3912
& 0.4779 & 0.3951
& 0.4708 & 0.3911
& 0.4824 & 0.4257
& \textbf{0.3225} & \textbf{0.3344}
\\

& 192
& \underline{0.5877} & \underline{0.4698}
& 0.6435 & 0.5066
& 0.7240 & 0.5107
& 0.6450 & 0.4971
& 0.8040 & 0.5385
& 0.7305 & 0.5194
& 0.7246 & 0.5501
& \textbf{0.5415} & \textbf{0.4443}
\\

\midrule

\multirow{3}{*}{
\rotatebox[origin=c]{90}{\textbf{\texttt{051201}}}
}
& 24
& 0.0090 & 0.0461
& 0.0093 & 0.0448
& \underline{0.0081} & \underline{0.0343}
& 0.0980 & 0.1883
& 0.0121 & 0.0551
& 0.0098 & 0.0401
& \underline{0.0081} & 0.0367
& \textbf{0.0072} & \textbf{0.0318}
\\

& 96
& 0.0555 & 0.1233
& 0.0569 & 0.1180
& 0.0516 & 0.0951
& 0.1394 & 0.2101
& 0.0565 & 0.1102
& 0.0530 & \underline{0.0930}
& \underline{0.0482} & 0.1009
& \textbf{0.0414} & \textbf{0.0867}
\\

& 192
& 0.1316 & 0.1981
& 0.1241 & 0.1884
& 0.1206 & \underline{0.1558}
& 0.2055 & 0.2689
& 0.1272 & 0.1702
& 0.1260 & 0.1593
& \underline{0.1130} & 0.1642
& \textbf{0.0992} & \textbf{0.1455}
\\

\bottomrule
\end{tabular}
\end{adjustbox}
\end{table*}

% \subsection{Computational Cost}
% % \begin{wrapfigure}{r}{0.45\columnwidth}
% %     \vspace{-1.2em}
% %     \centering
% %     \includegraphics[width=\linewidth]{figure/ICML-compt.pdf}
% %     \caption{Trade-offs among MSE, FLOPs, and memory on BJ.}
% %     \label{fig:efficiency}
% %     \vspace{-1.5em}
% % \end{wrapfigure}
% Fig.~\ref{fig:efficiency} compares MSE, FLOPs, and memory usage across models on the BJ dataset. Despite incorporating an image modality, XFMNet maintains moderate computational cost thanks to its lightweight visual encoder and compact fusion mechanism. Its FLOPs per iteration are comparable to mid-scale time-series models, and it uses 4.32\,GB of memory, lower than large multimodal or foundation models.
% Notably. 

% --------------------------------------------------------------------------

% --------------------------------------------------------------------------
\section{Conclusion}
\label{sec:conclusion}
In this work, we presented RaiNet, a rainfall-informed multimodal framework for long-term water-quality forecasting. RaiNet models multiscale nonstationary dynamics through \texttt{LocTrend}, converts gridded precipitation into basin-relative station-aligned rainfall events, and selectively integrates them using lag-aware \texttt{XGateFusion}. Experiments on three real-world datasets demonstrate consistent improvements over competitive forecasting and multimodal baselines, while qualitative analyses verify the distinct roles of each module. These results provide a data-driven approach toward accurate future water-quality forecasting.

\bibliography{iclr2027_conference}
\bibliographystyle{iclr2027_conference}

\appendix
\section{Appendix}

\subsection{Multiscale Embedding Strategy}
\subsubsection{Temporal (Sensor) Modality.}
\label{app:embedding}
At scale $l$, the water-quality sequence is denoted by
$\mathbf{X}^{(l)}\in\mathbb{R}^{T_l\times N}$, where $T_l$ and $N$ are the temporal length and number of monitoring stations, respectively. After scale-wise normalization, a convolutional value embedding is applied independently to each station sequence:
\begin{equation}
\mathbf{e}_t^{\mathrm{val}}
=
\operatorname{Conv1D}_3
\left(
\operatorname{Norm}_l(\mathbf{X}^{(l)})
\right)_t,
\qquad
\mathbf{e}_t^{\mathrm{time}}
=
\mathbf{m}_t\mathbf{W}_{\mathrm{time}},
\qquad
\mathbf{e}_t^{\mathrm{val}},
\mathbf{e}_t^{\mathrm{time}}
\in\mathbb{R}^{d},
\end{equation}
where $\operatorname{Conv1D}_3(\cdot)$ is a one-dimensional convolution with kernel size $3$ and circular padding, and $\operatorname{Norm}_l(\cdot)$ denotes normalization at scale $l$. The timestamp vector
$\mathbf{m}_t\in\mathbb{R}^{4}$ contains the normalized hour of day, day of week, day of month, and day of year at time step $t$, while
$\mathbf{W}_{\mathrm{time}}\in\mathbb{R}^{4\times d}$ is a learnable linear projection. The value and temporal-feature embeddings are summed to produce
\begin{equation}
\mathbf{F}_T^{(l)}
=
\left[
\operatorname{Dropout}
\left(
\mathbf{e}_t^{\mathrm{val}}
+
\mathbf{e}_t^{\mathrm{time}}
\right)
\right]_{t=1}^{T_l}
\in\mathbb{R}^{T_l\times d}.
\end{equation}
The embedding parameters are shared across temporal scales, and the timestamp features are subsampled consistently with the water-quality sequence at each scale.

\subsubsection{Rainfall-Event Encoder Details}
\label{app:rain_event}

\paragraph{Station-centered spatial prior.}
The spatial prior of station $n$ is constructed by evaluating a Gaussian kernel over the rainfall grid:
\begin{equation}
\pi_{n,p}
=
\frac{
\exp\!\left(
-\|\mathbf{z}_p-\mathbf{c}_n\|_2^2/(2\sigma^2)
\right)
}{
\sum_{p'}
\exp\!\left(
-\|\mathbf{z}_{p'}-\mathbf{c}_n\|_2^2/(2\sigma^2)
\right)
},
\end{equation}
where $\mathbf{z}_p$ is the coordinate of grid cell $p$, $\mathbf{c}_n$ is the geographic location of station $n$, and $\sigma$ controls the spatial extent of the station neighborhood. At each time step, invalid rainfall cells are excluded and the remaining prior weights are renormalized, ensuring that the station-local rainfall is computed only from valid observations.

\paragraph{Training-only event threshold.}
The station-specific threshold $\epsilon_n$ is fitted using only rainfall histories associated with the training origins:
\begin{equation}
\epsilon_n
=
\sqrt{
\frac{1}{|\mathcal{S}_{\mathrm{tr}}|}
\sum_{t\in\mathcal{S}_{\mathrm{tr}}}
\left(r_{n,t}^{\mathrm{rel}}\right)^2
},
\end{equation}
where $\mathcal{S}_{\mathrm{tr}}$ denotes the collection of historical rainfall time steps contained in the training samples. This root-mean-square threshold characterizes the typical magnitude of the local-to-basin contrast at each station. The fitted thresholds remain fixed during validation and testing.

\paragraph{Persistence aggregation and scale alignment.}
The time-indexed states $e_{n,t}$ are partitioned into non-overlapping
persistence blocks using $h=6$. Signed voting assigns one state to each
block, which is repeated over all positions in the corresponding block to
construct $\mathbf{F}_I\in\mathbb{R}^{T\times N}$. The block construction is
adapted to the sampling interval of each dataset. For scale $l$, the aligned
event sequence is computed as
%The hourly states $e_{n,t}$ are partitioned into non-overlapping blocks of length $h=6$. Signed voting assigns one state to each block, and this state is repeated over its six hourly positions to construct $\mathbf{F}_I\in\mathbb{R}^{T\times N}$. For scale $l$, the aligned event sequence is computed as
\begin{equation}
\big[\mathbf{F}_I^{(l)}\big]_{k,n}
=
\operatorname{sign}
\left(
\sum_{s=(k-1)q_l+1}^{kq_l}
\big[\mathbf{F}_I\big]_{s,n}
\right),
\qquad
k=1,\ldots,T_l,
\end{equation}
where $q_l$ is the aggregation factor shared with the corresponding water-quality scale, $k$ indexes the aligned temporal positions, and $T_l=T/q_l$ is the resulting sequence length. The summation accumulates event evidence over the temporal support of scale $l$, while the sign preserves its dominant local-to-basin direction. All blocks contain only historical rainfall observations available at the forecast origin.

\subsection{Detailed Derivation and Theoretical Guarantees for \texttt{LocTrend}}
\label{locccccc}

\subsubsection{Training-Only Structural Dictionary Fitting}
\label{app:loctrend_init}
The fitting procedure is performed before ordinary model training and uses neither validation nor test samples. It contains two passes: the first extracts temporal directions shared by centered local windows, and the second converts these directions into feature-space atoms used for background reconstruction. This two-pass construction separates \emph{when} recurrent structures occur within a local window from \emph{how} they are represented in the latent feature space: Pass 1 identifies shared temporal organizations, while Pass 2 converts them into reusable feature-space atoms. Let
$\mathcal{Q}_{\mathrm{tr}}$
denote the collection of overlapping windows extracted from all training origins and decomposed temporal scales. Each centered window is written as
$\widetilde{\mathcal{W}}_q
=[\widetilde{\mathbf{w}}_{q,1},\ldots,
\widetilde{\mathbf{w}}_{q,w}]
\in\mathbb{R}^{w\times d}$,
where $q\in\mathcal{Q}_{\mathrm{tr}}$ indexes a training window, $w$ is its length, and $d$ is the feature dimension.

\paragraph{Pass 1: shared temporal eigendirections.}
For each window, we first compute the pairwise squared distances
\begin{equation}
d_{q,t,u}^{2}
=
\left\|
\widetilde{\mathbf{w}}_{q,t}
-
\widetilde{\mathbf{w}}_{q,u}
\right\|_2^2,
\qquad t,u\in\{1,\ldots,w\}.
\end{equation}
The RBF bandwidth is adapted to the local feature scale using the median nonzero pairwise distance:
\begin{equation}
\gamma_q
=
\left(
\operatorname{median}_{t<u}
d_{q,t,u}^{2}
\right)^{-1}.
\end{equation}
Numerically, the denominator is lower-bounded by a small positive constant. The corresponding RBF Gram matrix is
\begin{equation}
[\mathbf{G}_q]_{t,u}
=
\exp\!\left(-\gamma_qd_{q,t,u}^{2}\right).
\end{equation}
Unlike a linear similarity matrix, $\mathbf{G}_q$ emphasizes nonlinear proximity between temporal positions whose feature states exhibit similar local structures. Each Gram matrix is centered as
\begin{equation}
\mathbf{G}_{q,c}
=
\mathbf{J}\mathbf{G}_q\mathbf{J},
\qquad
\mathbf{J}
=
\mathbf{I}_w-\frac{1}{w}\mathbf{1}_w\mathbf{1}_w^\top,
\end{equation}
where $\mathbf{I}_w$ is the $w\times w$ identity matrix and $\mathbf{1}_w$ is an all-one vector of length $w$. We then average the centered matrices over all training windows:
\begin{equation}
\overline{\mathbf{G}}
=
\frac{1}{|\mathcal{Q}_{\mathrm{tr}}|}
\sum_{q\in\mathcal{Q}_{\mathrm{tr}}}
\mathbf{G}_{q,c}.
\end{equation}
The top $K$ eigenpairs of this shared matrix satisfy
\begin{equation}
\overline{\mathbf{G}}\mathbf{u}_r
=
\lambda_r\mathbf{u}_r,
\qquad
r=1,\ldots,K,
\end{equation}
where $\lambda_r$ and $\mathbf{u}_r\in\mathbb{R}^{w}$ are the $r$-th retained eigenvalue and temporal eigendirection, respectively. Because they are extracted from the average kernel geometry, these directions summarize nonlinear temporal organizations recurring across the training windows instead of patterns from a single sample.

\paragraph{Pass 2: feature-space atom construction.}
The temporal eigendirections are next applied to every centered training window. For component $r$, we compute
\begin{equation}
\mathbf{z}_{q,r}
=
\frac{1}{\sqrt{\lambda_r}}
\sum_{t=1}^{w}
u_{t,r}\widetilde{\mathbf{w}}_{q,t}
\in\mathbb{R}^{d},
\qquad
\overline{\mathbf{z}}_{q,r}
=
\frac{\mathbf{z}_{q,r}}
{\|\mathbf{z}_{q,r}\|_2},
\end{equation}
where $u_{t,r}$ is element $t$ of $\mathbf{u}_r$. Thus, each temporal direction produces a normalized direction in the feature space of a training window. In other words, $u_{t,r}$ acts as a temporal weighting pattern over the window, projecting the $w$ temporal tokens into a $d$-dimensional feature direction $\mathbf{z}_{q,r}$. Components associated with larger kernel eigenvalues receive greater weight: $\omega_r
=
\frac{\lambda_r}{\max_{j}\lambda_j}.$
Their weighted feature covariance is accumulated over all windows:
\begin{equation}
\mathbf{C}_{F}
=
\frac{1}{K|\mathcal{Q}_{\mathrm{tr}}|}
\sum_{q\in\mathcal{Q}_{\mathrm{tr}}}
\sum_{r=1}^{K}
\omega_r
\overline{\mathbf{z}}_{q,r}
\overline{\mathbf{z}}_{q,r}^{\top}
\in\mathbb{R}^{d\times d}.
\end{equation}
Finally, the top $K$ eigenvectors of $\mathbf{C}_{F}$ form the structural dictionary:
\begin{equation}
\mathbf{C}_{F}\mathbf{k}_r
=
\xi_r\mathbf{k}_r,
\qquad
\mathbf{K}
=
\{\mathbf{k}_r\}_{r=1}^{K},
\end{equation}
where $\xi_r$ is the corresponding feature-covariance eigenvalue and
$\mathbf{k}_r\in\mathbb{R}^{d}$
is a structural atom. Eigenvector signs are fixed deterministically for reproducibility.

The same fitted dictionary is shared across all decomposed scales. In the default configuration, it remains fixed during ordinary training. During the LocTrend forward pass, it is symmetrically expanded as
$\mathbf{K}_{\pm}=[\mathbf{K};-\mathbf{K}]$,
and centered tokens are reconstructed through temperature-scaled cosine soft assignment. Therefore, the RBF kernel is used only to identify recurring training-set structures during dictionary fitting.

\subsubsection{LocTrend Forward Decomposition}
\label{app:loctrend_forward}
The procedure is applied independently to each station and each decomposed scale
$l=1,\ldots,L-L_{\mathrm{dir}}$. For scale $l$, the inputs are the temporal representation
$\mathbf{F}_T^{(l)}$,
the fitted dictionary
$\mathbf{K}=\{\mathbf{k}_r\}_{r=1}^{K}$,
the window length $w_l$, stride $s_l$, assignment temperature $\vartheta$, and reconstruction gain $\rho$. The outputs are the structural background
$\mathbf{B}^{(l)}$
and dynamic deviation
$\mathbf{D}^{(l)}$.

\paragraph{Window coverage.}
The window starting positions are collected in
\begin{equation}
\mathcal{Q}_l
=
\left\{
1,\,1+s_l,\,1+2s_l,\ldots
\right\}
\cup
\left\{T_l-w_l+1\right\},
\end{equation}
where only positions not exceeding $T_l-w_l+1$ are retained. The final position is included when necessary so that every temporal position is covered by at least one window. For a window beginning at $q_i\in\mathcal{Q}_l$, we extract and center
\begin{equation}
\mathcal{W}_i^{(l)}
=
\mathbf{F}_{T,\,q_i:q_i+w_l-1}^{(l)},
\qquad
\boldsymbol{\mu}_i^{(l)}
=
\frac{1}{w_l}
\sum_{t=1}^{w_l}
\mathcal{W}_{i,t}^{(l)},
\qquad
\widetilde{\mathbf{w}}_{i,t}^{(l)}
=
\mathcal{W}_{i,t}^{(l)}
-
\boldsymbol{\mu}_i^{(l)}.
\end{equation}
This removes the local level before structural matching, so dictionary assignment depends on local shape rather than absolute magnitude.

\paragraph{Signed soft reconstruction.}
The fitted dictionary is symmetrically expanded as
$\mathbf{K}_{\pm}=[\mathbf{K};-\mathbf{K}]$. The symmetric expansion allows the same structural atom to represent locally recurring patterns with opposite feature directions, avoiding the need to learn separate atoms for both signs.
For each centered token, the assignment weights are
\begin{equation}
a_{i,t,r}^{(l)}
=
\frac{
\exp\!\left(
\operatorname{cos}
\left(
\widetilde{\mathbf{w}}_{i,t}^{(l)},
\mathbf{k}_{\pm,r}
\right)/\vartheta
\right)
}{
\sum_{j=1}^{2K}
\exp\!\left(
\operatorname{cos}
\left(
\widetilde{\mathbf{w}}_{i,t}^{(l)},
\mathbf{k}_{\pm,j}
\right)/\vartheta
\right)
}.
\end{equation}
%The corresponding centered reconstruction and local background are
The resulting centered structural reconstruction is then combined with the local level to obtain the window-wise background:
\begin{equation}
\widetilde{\mathbf{b}}_{i,t}^{(l)}
=
\sum_{r=1}^{2K}
a_{i,t,r}^{(l)}
\mathbf{k}_{\pm,r},
\qquad
\mathbf{b}_{i,t}^{(l)}
=
\boldsymbol{\mu}_i^{(l)}
+
\rho\widetilde{\mathbf{b}}_{i,t}^{(l)}.
\end{equation}

\paragraph{Overlap aggregation.}
Each local reconstruction is added to its original temporal position. Let
$\Omega_\tau^{(l)}$
denote the set of windows covering global position $\tau$. The structural background and exact dynamic remainder are
\begin{equation}
\mathbf{B}_\tau^{(l)}
=
\frac{1}{|\Omega_\tau^{(l)}|}
\sum_{i\in\Omega_\tau^{(l)}}
\mathbf{b}_{i,\,\tau-q_i+1}^{(l)},
\qquad
\mathbf{D}^{(l)}
=
\mathbf{F}_T^{(l)}
-
\mathbf{B}^{(l)}.
\end{equation}
Coverage normalization prevents positions appearing in more overlapping windows from receiving systematically larger background values. The full forward procedure is summarized in Algorithm~\ref{alg:loctrend_forward}.

\begin{algorithm}[t]
\caption{\texttt{LocTrend} Forward Decomposition}
\label{alg:loctrend_forward}
\begin{algorithmic}[1]
    \STATE \textbf{Input:}
    temporal representation $\mathbf{F}_T^{(l)}$;
    fitted dictionary $\mathbf{K}$;
    window length $w_l$; stride $s_l$;
    temperature $\vartheta$; gain $\rho$
    
    \STATE \textbf{Output:}
    structural background $\mathbf{B}^{(l)}$;
    dynamic deviation $\mathbf{D}^{(l)}$

    \STATE Form the signed dictionary
    $\mathbf{K}_{\pm}\leftarrow[\mathbf{K};-\mathbf{K}]$

    \STATE Construct window starts $\mathcal{Q}_l$, including the final
    start $T_l-w_l+1$ when required

    \STATE Initialize
    $\mathbf{B}_{\mathrm{sum}}^{(l)}
    \leftarrow\mathbf{0}_{T_l\times d}$
    and coverage counts
    $\mathbf{m}^{(l)}\leftarrow\mathbf{0}_{T_l}$

    \FOR{each starting position $q_i\in\mathcal{Q}_l$}
        \STATE Extract
        $\mathcal{W}_i^{(l)}
        \leftarrow
        \mathbf{F}_{T,\,q_i:q_i+w_l-1}^{(l)}$

        \STATE Compute
        $\boldsymbol{\mu}_i^{(l)}
        \leftarrow
        w_l^{-1}
        \sum_{t=1}^{w_l}
        \mathcal{W}_{i,t}^{(l)}$

        \STATE Center
        $\widetilde{\mathcal{W}}_i^{(l)}
        \leftarrow
        \mathcal{W}_i^{(l)}
        -\boldsymbol{\mu}_i^{(l)}$

        \FOR{$t=1$ to $w_l$}
            \STATE Compute cosine similarities between
            $\widetilde{\mathbf{w}}_{i,t}^{(l)}$
            and all atoms
            $\mathbf{k}_{\pm,r}$ in $\mathbf{K}_{\pm}$

            \STATE Obtain assignment weights
            $\{a_{i,t,r}^{(l)}\}_{r=1}^{2K}
            \leftarrow
            \operatorname{softmax}_r
            \left(
            \operatorname{cos}
            \left(
            \widetilde{\mathbf{w}}_{i,t}^{(l)},
            \mathbf{k}_{\pm,r}
            \right)/\vartheta
            \right)$

            \STATE Reconstruct
            $\widetilde{\mathbf{b}}_{i,t}^{(l)}
            \leftarrow
            \sum_{r=1}^{2K}
            a_{i,t,r}^{(l)}
            \mathbf{k}_{\pm,r}$

            \STATE Restore the local level and scale the structural reconstruction:
            $\mathbf{b}_{i,t}^{(l)}
            \leftarrow
            \boldsymbol{\mu}_i^{(l)}
            +\rho\widetilde{\mathbf{b}}_{i,t}^{(l)}$

            \STATE Set the global position
            $\tau\leftarrow q_i+t-1$

            \STATE Accumulate the local reconstruction:
            $\mathbf{B}_{\mathrm{sum},\tau}^{(l)}
            \leftarrow
            \mathbf{B}_{\mathrm{sum},\tau}^{(l)}
            +\mathbf{b}_{i,t}^{(l)}$

            \STATE Update the coverage count:
            $m_\tau^{(l)}
            \leftarrow
            m_\tau^{(l)}+1$
        \ENDFOR
    \ENDFOR

    \STATE Normalize overlapping reconstructions:
    $\mathbf{B}_\tau^{(l)}
    \leftarrow
    \mathbf{B}_{\mathrm{sum},\tau}^{(l)}
    /m_\tau^{(l)}$,
    for $\tau=1,\ldots,T_l$

    \STATE Compute the exact dynamic deviation:
    $\mathbf{D}^{(l)}
    \leftarrow
    \mathbf{F}_T^{(l)}
    -\mathbf{B}^{(l)}$

    \STATE \textbf{return}
    $\mathbf{B}^{(l)},\mathbf{D}^{(l)}$
\end{algorithmic}
\end{algorithm}

\subsubsection{Local Adaptation of \texttt{LocTrend}}
\label{app:loctrend_theory}

We formalize how \texttt{LocTrend} adapts to locally shared level shifts in the temporal representation without relying on a predefined period. The scale and station indices are omitted for clarity. For temporal position $t$, let $\Omega_t$ denote the set of overlapping windows covering $t$, and define its effective local neighborhood as
\begin{equation}
\mathcal{N}_t
=
\bigcup_{i\in\Omega_t}
\left\{
q_i,\ldots,q_i+w-1
\right\},
\end{equation}
where $q_i$ and $w$ are the starting position and length of window $i$, respectively.

\begin{theorem}[Local Level Equivariance of \texttt{LocTrend}]
\label{thm:loctrend_local_equivariance}
Let $\mathbf{F}$ be an input representation with \texttt{LocTrend} decomposition
$\mathbf{F}=\mathbf{B}+\mathbf{D}$.
For any constant feature offset
$\boldsymbol{\delta}\in\mathbb{R}^{d}$, construct a locally shifted input
\begin{equation}
\mathbf{F}'_\tau
=
\begin{cases}
\mathbf{F}_\tau+\boldsymbol{\delta},
& \tau\in\mathcal{N}_t,\\
\mathbf{F}_\tau,
& \tau\notin\mathcal{N}_t.
\end{cases}
\end{equation}
Then the \texttt{LocTrend} decomposition at position $t$ satisfies
\begin{equation}
\mathbf{B}'_t
=
\mathbf{B}_t+\boldsymbol{\delta},
\qquad
\mathbf{D}'_t
=
\mathbf{D}_t.
\end{equation}
\end{theorem}

\begin{proof}
For every window $i\in\Omega_t$, all tokens in that window belong to
$\mathcal{N}_t$ and are shifted by the same vector $\boldsymbol{\delta}$.
Its local mean therefore becomes
\begin{equation}
\boldsymbol{\mu}'_i
=
\boldsymbol{\mu}_i+\boldsymbol{\delta}.
\end{equation}
For every local position $j=1,\ldots,w$, the centered token remains unchanged:
\begin{equation}
\widetilde{\mathbf{w}}'_{i,j}
=
\left(\mathbf{w}_{i,j}+\boldsymbol{\delta}\right)
-
\left(\boldsymbol{\mu}_i+\boldsymbol{\delta}\right)
=
\widetilde{\mathbf{w}}_{i,j}.
\end{equation}
Consequently, its cosine similarities to the signed dictionary, soft assignments, and centered reconstruction remain unchanged. Restoring the shifted local mean therefore gives
\begin{equation}
\mathbf{b}'_{i,j}
=
\mathbf{b}_{i,j}
+
\boldsymbol{\delta}.
\end{equation}
Coverage-normalized averaging over all windows in $\Omega_t$ yields
\begin{equation}
\mathbf{B}'_t
=
\frac{1}{|\Omega_t|}
\sum_{i\in\Omega_t}
\mathbf{b}'_{i,\,t-q_i+1}
=
\mathbf{B}_t+\boldsymbol{\delta}.
\end{equation}
Since
$\mathbf{F}'_t=\mathbf{F}_t+\boldsymbol{\delta}$
and
$\mathbf{D}_t=\mathbf{F}_t-\mathbf{B}_t$,
we obtain
\begin{equation}
\mathbf{D}'_t
=
\mathbf{F}'_t-\mathbf{B}'_t
=
\mathbf{F}_t-\mathbf{B}_t
=
\mathbf{D}_t.
\end{equation}
\end{proof}

The theorem shows that \texttt{LocTrend} absorbs a locally shared representation-level shift into the structural background while leaving the dynamic deviation unchanged. This property follows directly from local centering and overlap reconstruction and does not require the shift to recur at a predefined temporal period.

\subsection{Component-Specific Cross-Scale Mixing}
\label{app:cross_scale_mixing}

Let $L_{\mathrm{dec}}=L-L_{\mathrm{dir}}$ denote the number of scales processed by \texttt{LocTrend}. Cross-scale mixing is performed only among these decomposed scales.

\begin{figure*}[htbp]
    \centering
    \includegraphics[width=\columnwidth]{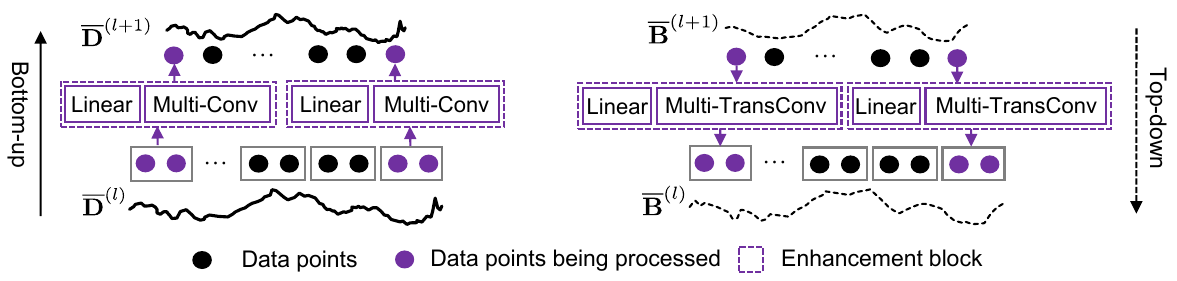}
    \caption{Component-specific cross-scale mixing with opposite propagation directions. Dynamic deviations are enhanced bottom-up from fine to coarse scales, whereas structural backgrounds are enhanced top-down from coarse to fine scales. At each transition, a temporal linear path and a multi-kernel convolutional or transposed-convolutional path are adaptively combined through learnable softmax fusion.}
    \label{fig:component_mixing}
\end{figure*}

\paragraph{Fine-to-coarse deviation mixing.}
Dynamic deviations are propagated sequentially from scale $l$ to the coarser scale $l+1$. We initialize
$\overline{\mathbf{D}}^{(1)}=\mathbf{D}^{(1)}$ and, for
$l=1,\ldots,L_{\mathrm{dec}}-1$, compute
\begin{align}
\mathbf{U}_{D,\mathrm{conv}}^{(l\rightarrow l+1)}
&=
\operatorname{Conv}_{\{3,5,7\},\downarrow}^{(l)}
\left(\overline{\mathbf{D}}^{(l)}\right),\\
\mathbf{U}_{D,\mathrm{lin}}^{(l\rightarrow l+1)}
&=
\operatorname{MLP}_{\downarrow}^{(l)}
\left(\overline{\mathbf{D}}^{(l)}\right),\\
\overline{\mathbf{D}}^{(l+1)}
&=
\mathbf{D}^{(l+1)}
+
\operatorname{SF}_{D}^{(l)}
\left(
\mathbf{U}_{D,\mathrm{conv}}^{(l\rightarrow l+1)},
\mathbf{U}_{D,\mathrm{lin}}^{(l\rightarrow l+1)}
\right).
\label{eq:appendix_deviation_mixing}
\end{align}
where $\operatorname{Conv}_{\{3,5,7\},\downarrow}^{(l)}$ denotes a strided temporal convolutional path containing kernels of sizes $3$, $5$, and $7$, while $\operatorname{MLP}_{\downarrow}^{(l)}$ maps the temporal length from $T_l$ to $T_{l+1}$. The former extracts local variations over multiple receptive fields, whereas the latter preserves a direct global transformation between the two scales.

\paragraph{Coarse-to-fine background mixing.}
Structural backgrounds follow the reverse direction. Starting from
$\overline{\mathbf{B}}^{(L_{\mathrm{dec}})}
=\mathbf{B}^{(L_{\mathrm{dec}})}$, we compute, for
$l=L_{\mathrm{dec}}-1,\ldots,1$,
\begin{align}
\mathbf{U}_{B,\mathrm{conv}}^{(l+1\rightarrow l)}
&=
\operatorname{TConv}_{\{3,5,7\},\uparrow}^{(l)}
\left(\overline{\mathbf{B}}^{(l+1)}\right),\\
\mathbf{U}_{B,\mathrm{lin}}^{(l+1\rightarrow l)}
&=
\operatorname{MLP}_{\uparrow}^{(l)}
\left(\overline{\mathbf{B}}^{(l+1)}\right),\\
\overline{\mathbf{B}}^{(l)}
&=
\mathbf{B}^{(l)}
+
\operatorname{SF}_{B}^{(l)}
\left(
\mathbf{U}_{B,\mathrm{conv}}^{(l+1\rightarrow l)},
\mathbf{U}_{B,\mathrm{lin}}^{(l+1\rightarrow l)}
\right).
\label{eq:appendix_background_mixing}
\end{align}
where $\operatorname{TConv}_{\{3,5,7\},\uparrow}^{(l)}$ denotes the multi-kernel transposed-convolutional path, and $\operatorname{MLP}_{\uparrow}^{(l)}$ expands the temporal length from $T_{l+1}$ to $T_l$.

For either direction, the two parallel paths are combined by learnable softmax fusion:
\begin{equation}
\operatorname{SF}
\left(\mathbf{U}_{\mathrm{conv}},\mathbf{U}_{\mathrm{lin}}\right)
=
\alpha_{\mathrm{conv}}\mathbf{U}_{\mathrm{conv}}
+
\alpha_{\mathrm{lin}}\mathbf{U}_{\mathrm{lin}},
\qquad
[\alpha_{\mathrm{conv}},\alpha_{\mathrm{lin}}]
=
\operatorname{softmax}(\boldsymbol{\gamma}),
\label{eq:softmax_path_fusion}
\end{equation}
where $\boldsymbol{\gamma}\in\mathbb{R}^{2}$ contains learnable path logits. This allows each transition to determine the relative contributions of local multi-receptive-field patterns and global temporal transformations. Finally, the enhanced components are recombined with the original temporal representation:
\begin{equation}
\widetilde{\mathbf{F}}_{T}^{(l)}
=
\mathbf{F}_{T}^{(l)}
+
\operatorname{FFN}
\left(
\overline{\mathbf{B}}^{(l)}
+
\overline{\mathbf{D}}^{(l)}
\right),
\qquad
l=1,\ldots,L_{\mathrm{dec}},
\label{eq:appendix_component_reconstruction}
\end{equation}
where $\operatorname{FFN}(\cdot)$ denotes the feed-forward transformation within the residual update. The direct scales $l>L_{\mathrm{dec}}$ bypass this operation and preserve
$\widetilde{\mathbf{F}}_{T}^{(l)}=\mathbf{F}_{T}^{(l)}$.

\subsection{FFT Lag Interaction and Dynamic Gate Construction}
\label{app:xgate}

For scale $l$ and station $n$, \texttt{XGateFusion} receives the baseline and rainfall correction
$\widehat{\mathbf{y}}_{T,n}^{(l)},
\widehat{\mathbf{y}}_{I,n}^{(l)}\in\mathbb{R}^{H}$.
It constructs
\begin{equation}
\mathbf{u}_{T,n}^{(l)}
=
\widehat{\mathbf{y}}_{T,n}^{(l)},
\qquad
\mathbf{u}_{I,n}^{(l)}
=
\widehat{\mathbf{y}}_{T,n}^{(l)}
+
\widehat{\mathbf{y}}_{I,n}^{(l)},
\label{eq:app_xgate_inputs}
\end{equation}
so that the interaction is evaluated between the baseline and its rainfall-conditioned counterpart. Independent input projections first produce
\begin{equation}
\mathbf{H}_{T,n}^{(l)}
=
\phi_T\!\left(\mathbf{u}_{T,n}^{(l)}\right),
\qquad
\mathbf{H}_{I,n}^{(l)}
=
\phi_I\!\left(\mathbf{u}_{I,n}^{(l)}\right)
\in\mathbb{R}^{H\times d},
\label{eq:app_xgate_projection}
\end{equation}
where $\phi_T$ and $\phi_I$ are learnable projections and $d$ is the hidden dimension.

\paragraph{Bidirectional aperiodic lag interaction.} For readability, the station index \(n\) is omitted within the lag-interaction equations.
For the full relative-lag domain
$\mathcal{T}_H=\{-(H-1),\ldots,H-1\}$, the two directional cross-correlation scores are
\begin{align}
\mathbf{C}_{T\leftarrow I}^{(l)}(\tau)
&=
\sum_t
\mathbf{Q}_{T}^{(l)}(t)
\odot
\mathbf{K}_{I}^{(l)}(t-\tau),\\
\mathbf{C}_{I\leftarrow T}^{(l)}(\tau)
&=
\sum_t
\mathbf{Q}_{I}^{(l)}(t)
\odot
\mathbf{K}_{T}^{(l)}(t-\tau),
\qquad \tau\in\mathcal{T}_H,
\label{eq:app_lag_correlations}
\end{align}
where $\mathbf{Q}$ and $\mathbf{K}$ denote query and key projections, and positions outside the forecast horizon are zero padded. Using an FFT length of at least $2H-1$ implements aperiodic linear correlation and avoids circular wraparound. The correlation scores are converted into bounded signed lag weights:
\begin{equation}
\mathbf{A}_{a\leftarrow b}^{(l)}(\tau)
=
\operatorname{Dropout}\!\left[
d_h^{-1/2}
\tanh\!\left(
\mathbf{C}_{a\leftarrow b}^{(l)}(\tau)
\right)
\right],
\label{eq:app_lag_weights}
\end{equation}
where $(a,b)\in\{(T,I),(I,T)\}$ and $d_h$ is the dimension of each attention head. The corresponding values are aggregated through linear lag convolution:
\begin{equation}
\mathbf{O}_{a\leftarrow b}^{(l)}(t)
=
\sum_{\tau\in\mathcal{T}_H}
\mathbf{A}_{a\leftarrow b}^{(l)}(\tau)
\odot
\mathbf{V}_{b}^{(l)}(t-\tau),
\label{eq:app_lag_aggregation}
\end{equation}
where $\mathbf{V}_{b}^{(l)}$ denotes the value projection of branch $b$. Both correlation and aggregation are evaluated by FFT with temporal complexity
$\mathcal{O}(H\log H)$ \cite{wu2021autoformer}. The two directions are combined symmetrically:
\begin{equation}
\mathbf{H}_{X,n}^{(l)}
=
\frac{1}{2}
\left[
\psi\!\left(
\mathbf{O}_{T\leftarrow I,n}^{(l)}
\right)
+
\psi\!\left(
\mathbf{O}_{I\leftarrow T,n}^{(l)}
\right)
\right],
\label{eq:app_xgate_interaction}
\end{equation}
where $\psi$ is the shared output projection. The relative lags describe learned statistical offsets between the two prediction sequences and should not be interpreted as causal rainfall-response delays.

\paragraph{Dynamic gate construction.}
Each scale has a learned scalar base gate $g^{(l)}$, which represents its overall reliance on the rainfall correction. The interaction representation further produces a bounded station and horizon-dependent update:
\begin{equation}
\Delta\mathbf{g}_{n}^{(l)}
=
\delta_{\max}
\tanh\!\left[
W_o\operatorname{LN}
\left(\mathbf{H}_{X,n}^{(l)}\right)
\right]
\in\mathbb{R}^{H},
\label{eq:app_dynamic_gate}
\end{equation}
where $\operatorname{LN}(\cdot)$ denotes layer normalization, $W_o$ is a learnable output projection, and $\delta_{\max}$ bounds the update magnitude. Collecting all station-wise updates gives
$\Delta\mathbf{G}^{(l)}
=[\Delta\mathbf{g}_{1}^{(l)},\ldots,
\Delta\mathbf{g}_{N}^{(l)}]
\in\mathbb{R}^{H\times N}$. The scale-wise fused prediction is therefore
\begin{equation}
\widehat{\mathbf{Y}}_{TI}^{(l)}
=
\widehat{\mathbf{Y}}_{T}^{(l)}
+
\left(
g^{(l)}\mathbf{1}_{H\times N}
+
\Delta\mathbf{G}^{(l)}
\right)
\odot
\widehat{\mathbf{Y}}_{I}^{(l)}.
\label{eq:app_xgate_output}
\end{equation}
The projection producing $\Delta\mathbf{G}^{(l)}$ is initialized to zero, making the initial output
$\widehat{\mathbf{Y}}_{T}^{(l)}
+g^{(l)}\widehat{\mathbf{Y}}_{I}^{(l)}$.
The \texttt{XGateFusion} modules follow the same architecture across scales but use independent parameters.

\subsection{Dataset Details}
\label{sec:appendix_dataset}

\paragraph{Data construction.}
We construct three multimodal datasets from USGS monitoring stations in basins 170900, 071200, and 051201\footnote{\url{https://waterdata.usgs.gov/nwis/}}.
Each dataset comprises monitoring stations within a shared drainage basin that are connected through the river network and provide concurrent records over the study period. The selected networks contain 15, 5, and 5 stations, respectively. Short missing segments are forward-filled, while backward filling is used only for missing values at the beginning of a sequence.

Table~\ref{tab:dataset_statistics} summarizes the resulting datasets. We study turbidity and nitrate because they represent complementary rainfall-related processes: turbidity reflects rapid sediment mobilization and runoff, whereas nitrate captures nutrient transport and leaching over broader timescales. Forecasting turbidity provides early indication of runoff-driven particle loads that can challenge water-treatment operations, while nitrate forecasting supports nutrient-pollution surveillance, source-water management, and eutrophication risk assessment. Basin 170900 provides five years of hourly turbidity observations, while basins 071200 and 051201 provide three and five years of nitrate observations, respectively. The 15-minute records of 051201 are aggregated to 30-minute intervals to match the native temporal resolution of the adopted IMERG product and avoid repeatedly pairing multiple water-quality observations with the same precipitation grid. Together, the datasets comprise 157,824 aligned multimodal time steps.

\begin{table*}[htbp]
\centering
\caption{Statistics of the three multimodal water-quality datasets.}
\label{tab:dataset_statistics}
{\fontsize{8pt}{10pt}\selectfont
\setlength{\tabcolsep}{8pt}
\renewcommand{\arraystretch}{1.08}
\begin{tabular}{lcccccc}
\toprule
\textbf{Basin} & \textbf{Target} & \textbf{Period} & \textbf{Interval}
& \textbf{Stations} & \textbf{Paired steps} & \textbf{Raster size} \\
\midrule
170900 & Turbidity & 2020/10--2025/09 & 1 h
& 15 & 43,824 & $7\times18$ \\
071200 & Nitrate & 2022/04--2025/03 & 1 h
& 5 & 26,304 & $8\times31$ \\
051201 & Nitrate & 2019/10--2024/09 & 30 min
& 5 & 87,696 & $37\times9$ \\
%\midrule
%\multicolumn{5}{r}{\textbf{Total}} & \textbf{157,824} & \\
\bottomrule
\end{tabular}}
\end{table*}

\begin{figure}[t]
\centering
\includegraphics[width=0.8\columnwidth]{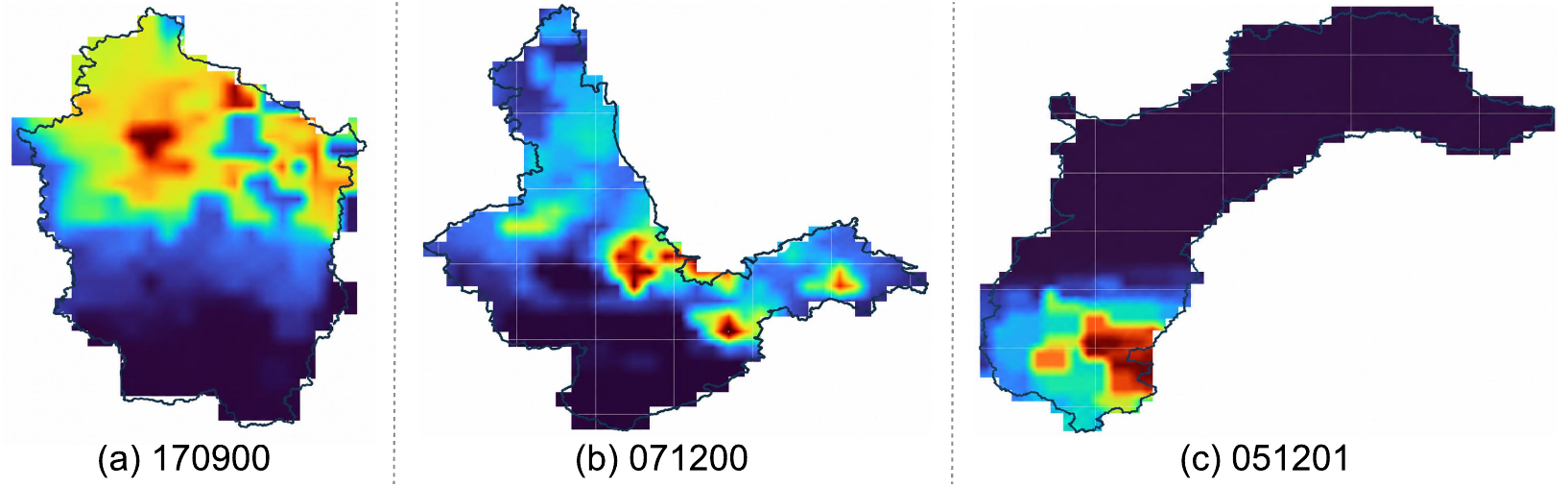}
\caption{Example of gridded IMERG precipitation.}
\label{rainex}
\end{figure}

\paragraph{Gridded precipitation.}
Precipitation fields are obtained from NASA GPM IMERG through Google Earth Engine.\footnote{\url{https://developers.google.com/earth-engine/datasets/catalog/NASA_GPM_L3_IMERG_V07}}
Each single-band GeoTIFF records precipitation intensity in
$\mathrm{mm\,h^{-1}}$ on a $0.1^\circ\times0.1^\circ$ grid. The raster extent is derived from the monitoring-station boundary with additional spatial padding, preserving both station-local rainfall and its surrounding basin context. The resulting spatial coverages are
$123.6$-$121.8^\circ$W and $45.0$-$45.7^\circ$N for 170900,
$89.2$-$86.1^\circ$W and $41.0$-$41.8^\circ$N for 071200, and
$88.2$-$87.4^\circ$W and $37.4$-$41.0^\circ$N for 051201.
Figure~\ref{rainex} presents an example of the gridded precipitation input.

\paragraph{Multimodal alignment.}
Water-quality observations and precipitation grids are matched by UTC timestamp. Hourly water-quality records are paired with hourly precipitation grids, while basin 051201 is evaluated at the native 30-minute IMERG interval. This produces a one-to-one correspondence between the two modalities. Table~\ref{tab:data_sample} provides 12 consecutive hourly nitrate observations from the five stations in basin 071200 as an example.

\begin{table*}[t]
\centering
\caption{Twelve-hour nitrate observations from five stations in basin 071200
($\mathrm{mg\,L^{-1}}$).}
\label{tab:data_sample}
{\fontsize{8pt}{10pt}\selectfont
\setlength{\tabcolsep}{7pt}
\renewcommand{\arraystretch}{1.0}
\begin{tabular}{lccccc}
\toprule
\textbf{Time (UTC)}
& \textbf{05515500}
& \textbf{05518000}
& \textbf{05536356}
& \textbf{05537980}
& \textbf{05543010} \\
\midrule
\textbf{Latitude}
& $41.390^\circ$N & $41.183^\circ$N & $41.619^\circ$N
& $41.536^\circ$N & $41.300^\circ$N \\
\textbf{Longitude}
& $86.706^\circ$W & $87.340^\circ$W & $87.500^\circ$W
& $88.084^\circ$W & $88.614^\circ$W \\
\midrule
2022/04/01 00:00 & 1.720 & 2.135 & 5.400 & 4.500 & 4.200 \\
2022/04/01 01:00 & 1.750 & 2.135 & 5.400 & 4.350 & 4.200 \\
2022/04/01 02:00 & 1.765 & 2.120 & 5.400 & 4.300 & 4.200 \\
2022/04/01 03:00 & 1.795 & 2.120 & 5.400 & 4.300 & 4.200 \\
2022/04/01 04:00 & 1.840 & 2.125 & 5.350 & 4.200 & 4.200 \\
2022/04/01 05:00 & 1.900 & 2.135 & 5.350 & 4.150 & 4.400 \\
2022/04/01 06:00 & 1.955 & 2.140 & 5.600 & 4.250 & 4.400 \\
2022/04/01 07:00 & 1.995 & 2.135 & 5.700 & 4.350 & 4.400 \\
2022/04/01 08:00 & 2.015 & 2.120 & 5.800 & 4.400 & 4.400 \\
2022/04/01 09:00 & 2.065 & 2.130 & 5.800 & 4.400 & 4.500 \\
2022/04/01 10:00 & 2.095 & 2.130 & 5.800 & 4.300 & 4.500 \\
2022/04/01 11:00 & 2.100 & 2.130 & 5.900 & 3.900 & 4.500 \\
\bottomrule
\end{tabular}}
\end{table*}

%\subsection{Hyperparameter Sensitivity} \label{sen}

\subsection{Hyperparameter Sensitivity}
\label{sen}
We evaluate the sensitivity of RaiNet to four parameters associated with its
main designs: the number of temporal scales $L$, rainfall-event persistence
setting $h$, LocTrend window configuration $(w_l,s_l)$, and dictionary size
$K$. We vary one parameter at a time while keeping the remaining settings at
their default values. 
%In particular, $q=2$, $L_{\mathrm{dir}}=1$, $\vartheta=1$, and $\rho=1$ remain fixed. 
For $(w_l,s_l)$, the stride is set
to half of the window length to maintain a consistent overlap ratio. The same default configuration is used for all three datasets without dataset-specific tuning. The number of scales $L$ controls the temporal support available to RaiNet, whereas $h$ determines how consecutive station-relative rainfall states are consolidated into persistent events. Table~\ref{tab:sensitivity_multiscale} evaluates
$L\in\{2,3,4,5\}$ and $h\in\{3,6,9,12\}$. These experiments examine the
balance between multiscale coverage and temporal detail, as well as between
suppressing isolated rainfall fluctuations and retaining event timing.

%Table~\ref{tab:sensitivity_multiscale} evaluates $L\in\{2,3,4,5\}$ and $h\in\{3,6,9,12\}$ sampling steps. These experiments examine the balance between multiscale coverage and temporal detail, as well as between suppressing isolated rainfall fluctuations and retaining event timing.

% =========================================================
% Table 1: Temporal scales and rainfall-event persistence
% =========================================================
\begin{table*}[htbp]
\centering
\nomakegapedcells
{\fontsize{7.4pt}{10pt}\selectfont
\setlength{\tabcolsep}{5.5pt}
\renewcommand{\arraystretch}{0.9}
\caption{Sensitivity to the number of temporal scales $L$ and rainfall-event
persistence setting $h$. Default settings are shown in bold in the column
headers.}
\label{tab:sensitivity_multiscale}
\begin{tabular}{lcccccccc}
\toprule
\multicolumn{1}{c}{\textbf{Dataset}} &
\multicolumn{2}{c}{$L=2$} &
\multicolumn{2}{c}{$L=3$} &
\multicolumn{2}{c}{$\mathbf{L=4}$} &
\multicolumn{2}{c}{$L=5$} \\
\cmidrule(l{7pt}r{7pt}){1-1}
\cmidrule(lr{7pt}){2-3}
\cmidrule(lr{7pt}){4-5}
\cmidrule(lr{7pt}){6-7}
\cmidrule(lr{7pt}){8-9}
\textbf{Metric} &
MSE & MAE & MSE & MAE & MSE & MAE & MSE & MAE \\
\midrule
170900 &
0.4337 & 0.2573 &
0.4711 & 0.2971 &
0.4067 & 0.2578 &
0.4144 & 0.2575 \\

071200 &
0.3255 & 0.3373 &
0.3279 & 0.3347 &
0.3238 & 0.3338 &
0.3275 & 0.3347 \\

051201 &
0.0439 & 0.0901 &
0.0417 & 0.0857 &
0.0422 & 0.0868 &
0.0412 & 0.0868 \\
\midrule
\multicolumn{1}{c}{\textbf{Dataset}} &
\multicolumn{2}{c}{$h=3$} &
\multicolumn{2}{c}{$\mathbf{h=6}$} &
\multicolumn{2}{c}{$h=9$} &
\multicolumn{2}{c}{$h=12$} \\
\cmidrule(l{7pt}r{7pt}){1-1}
\cmidrule(lr{7pt}){2-3}
\cmidrule(lr{7pt}){4-5}
\cmidrule(lr{7pt}){6-7}
\cmidrule(lr{7pt}){8-9}
\textbf{Metric} &
MSE & MAE & MSE & MAE & MSE & MAE & MSE & MAE \\
\midrule
170900 &
0.4129 & 0.2638 &
0.4066 & 0.2572 &
0.4106 & 0.2561 &
0.4240 & 0.2675 \\

071200 &
0.3274 & 0.3355 &
0.3227 & 0.3330 &
0.3159 & 0.3310 &
0.3296 & 0.3357 \\

051201 &
0.0416 & 0.0863 &
0.0417 & 0.0866 &
0.0412 & 0.0867 &
0.0422 & 0.0869 \\
\bottomrule
\end{tabular}
}
\end{table*}

The results show that a moderate number of temporal scales is sufficient for
capturing multiscale water-quality dynamics. In particular, $L=4$ achieves
the lowest MSE on 170900 and both the lowest MSE and MAE on 071200, while
remaining close to the best results on 051201. Increasing the number of scales
further does not consistently improve performance, suggesting that excessive
downsampling provides limited additional benefit once sufficiently broad
temporal supports have been covered. For rainfall-event persistence, moderate values generally perform better than
an overly long aggregation interval. The strongest results are mainly observed
around $h=6$-$9$, whereas $h=12$ tends to degrade performance. This behavior is consistent with the role of persistence aggregation: short aggregation intervals retain transient fluctuations, while excessively long intervals may merge temporally distinct rainfall events and blur their timing. We therefore retain $h=6$ as a common short-range persistence setting across datasets rather
than tuning the event duration separately for each dataset. We set $L_{\mathrm{dir}}=1$, so the last coarsest scale follows the direct
temporal path. As shown in Fig.~\ref{locloc}, the structural background becomes nearly flat by the third decomposed scale, indicating that further decomposition after additional downsampling provides limited structural benefit. We preserve the fourth and coarsest scale without further decomposition.

% =========================================================
% Table 2: LocTrend window and dictionary size
% =========================================================
\begin{table*}[htbp]
\centering
\nomakegapedcells
{\fontsize{7.4pt}{10pt}\selectfont
\setlength{\tabcolsep}{5.5pt}
\renewcommand{\arraystretch}{0.9}
\caption{Sensitivity to the LocTrend window configuration $(w_l,s_l)$ and
shared dictionary size $K$. Default settings are shown in bold in the column
headers.}
\label{tab:sensitivity_loctrend}
\begin{tabular}{lcccccccccc}
\toprule
\multicolumn{1}{c}{\textbf{Dataset}} &
\multicolumn{2}{c}{$(12,6)$} &
\multicolumn{2}{c}{$(18,9)$} &
\multicolumn{2}{c}{$\mathbf{(24,12)}$} &
\multicolumn{2}{c}{$(30,15)$} &
\multicolumn{2}{c}{$(36,18)$} \\
\cmidrule(l{7pt}r{7pt}){1-1}
\cmidrule(lr{7pt}){2-3}
\cmidrule(lr{7pt}){4-5}
\cmidrule(lr{7pt}){6-7}
\cmidrule(lr{7pt}){8-9}
\cmidrule(lr{7pt}){10-11}
\textbf{Metric} &
MSE & MAE & MSE & MAE & MSE & MAE &
MSE & MAE & MSE & MAE \\
\midrule
170900 &
0.4060 & 0.2542 &
0.4213 & 0.2680 &
0.4092 & 0.2599 &
0.4149 & 0.2653 &
0.4590 & 0.2940 \\

071200 &
0.3227 & 0.3331 &
0.3190 & 0.3334 &
0.3219 & 0.3332 &
0.3223 & 0.3318 &
0.3269 & 0.3340 \\

051201 &
0.0406 & 0.0855 &
0.0422 & 0.0869 &
0.0411 & 0.0857 &
0.0420 & 0.0861 &
0.0417 & 0.0861 \\
\midrule
\multicolumn{1}{c}{\textbf{Dataset}} &
\multicolumn{2}{c}{$K=1$} &
\multicolumn{2}{c}{$K=2$} &
\multicolumn{2}{c}{$\mathbf{K=4}$} &
\multicolumn{2}{c}{$K=6$} &
\multicolumn{2}{c}{$K=8$} \\
\cmidrule(l{7pt}r{7pt}){1-1}
\cmidrule(lr{7pt}){2-3}
\cmidrule(lr{7pt}){4-5}
\cmidrule(lr{7pt}){6-7}
\cmidrule(lr{7pt}){8-9}
\cmidrule(lr{7pt}){10-11}
\textbf{Metric} &
MSE & MAE & MSE & MAE & MSE & MAE &
MSE & MAE & MSE & MAE \\
\midrule
170900 &
0.4025 & 0.2567 &
0.4008 & 0.2541 &
0.4072 & 0.2609 &
0.4076 & 0.2646 &
0.4051 & 0.2606 \\

071200 &
0.3232 & 0.3336 &
0.3235 & 0.3332 &
0.3243 & 0.3343 &
0.3242 & 0.3340 &
0.3219 & 0.3345 \\

051201 &
0.0416 & 0.0869 &
0.0414 & 0.0868 &
0.0422 & 0.0869 &
0.0415 & 0.0860 &
0.0410 & 0.0855 \\
\bottomrule
\end{tabular}
}
\end{table*}

Table~\ref{tab:sensitivity_loctrend} shows that LocTrend remains reasonably
stable over a range of moderate window configurations. Although the exact
optimum varies across datasets, $(24,12)$ remains close to the best-performing
settings in all three cases. In contrast, excessively long windows can reduce
accuracy, as most clearly observed for $(36,18)$ on 170900. This suggests that
overly broad temporal neighborhoods may weaken the local adaptivity required
for background reconstruction. We therefore use $(24,12)$ as a common
moderate-size configuration across datasets. The effect of dictionary size is similarly non-monotonic. Increasing $K$ does
not consistently improve either MSE or MAE, and different datasets favor
different dictionary sizes. This indicates that LocTrend does not rely on a
large structural dictionary to obtain accurate forecasts; a small number of
shared atoms is already sufficient to represent the recurring structures
needed for background reconstruction. We therefore use $K=4$ as a fixed
compact dictionary size throughout the experiments.

The sensitivity results show that RaiNet does not depend on a single
dataset-specific hyperparameter optimum. The adopted configuration,
$L=4$, $h=6$, $(w_l,s_l)=(24,12)$, and $K=4$, provides a consistent operating
point across the three datasets and is kept unchanged throughout the main
experiments.

\subsection{Comparative Models}
\label{cm}
We compare RaiNet with four groups of representative forecasting models. All baselines follow the same chronological data splits, input/output horizons, normalization, and evaluation metrics as RaiNet. 
%When an original model is not directly compatible with our multi-station water-quality setting, we preserve its core modeling mechanism and make only task-oriented adaptations to match the forecasting targets.

\paragraph{Time-series models.}
TimeKAN \citep{TimeKAN} models multiscale temporal patterns through frequency decomposition and Kolmogorov-Arnold Networks. 
FilterTS \citep{FilterTS} learns adaptive frequency-domain filters for temporal forecasting. 
GTR \citep{cao2026gtr} retrieves informative global patterns beyond the local input window, while MixLinear \citep{ma2026mixlinear} combines temporal and frequency-domain linear modeling. They represent complementary temporal modeling paradigms spanning multiscale decomposition, adaptive filtering, global retrieval, and dual-domain linear modeling. 

\paragraph{Spatio-temporal models.}
HyperD \citep{shao2026hyperd} decouples periodic and residual dynamics for spatio-temporal forecasting. 
RSTIB-MLP \citep{chen2025information} introduces an information-bottleneck objective for robust spatio-temporal representation learning. 
ST-SSDL \citep{gao2026different} explicitly models deviations from historical patterns through self-supervised learning. 
These methods represent complementary paradigms of periodicity decoupling, robust dependency learning, and deviation-aware modeling.

\paragraph{Water-quality-specific models.}
PIWON \citep{tang2026non} explicitly models pollutant accumulation and wash-off processes. 
We retain its original accumulation and wash-off recurrence, but replace unavailable event-level rainfall-runoff inputs with historical water quality, rainfall statistics, and discharge, and add a multi-station multi-step prediction head. D-TGCN \citep{jiang2024heterogeneous} models directed dependencies among monitoring stations according to river-network topology.
CMLIP \citep{bi2025hybrid} jointly models water-quality observations and rainfall information. 
In our independent implementation, its original water-quality-variable formulation is adapted to multi-station forecasting, while small rainfall rasters are zero-padded to satisfy the visual encoder input requirement. RGCN \citep{smith2024predictive} combines graph and recurrent modeling for stream-quality prediction. 
We preserve its gated graph-recurrent backbone, remove unavailable distance, conductivity, baseflow, land-cover, road-salt, and meteorological variables, use historical water quality with upstream-downstream topology, and add a multi-step prediction head. XGBR \citep{ali2025near} is adapted from next-day \textit{E. coli} prediction to recursive multi-step turbidity/nitrate forecasting using the available water-quality, discharge, rainfall, and calendar features. They cover diverse water-quality-specific paradigms, including physics-informed, graph-based, multimodal, recurrent, and tree-based modeling.

\paragraph{Diffusion-based models.}
SimDiff \citep{ding2026simdiff} formulates time-series forecasting through an end-to-end diffusion process, directly generating future trajectories from noisy states. 
MA-TSD \citep{wang2025non} introduces moving-average-guided diffusion transitions to better accommodate nonstationary temporal dynamics. They represent complementary diffusion forecasting paradigms based on direct generative prediction and structured nonstationarity modeling.

\subsection{Full Comparative Result and Computational Cost}
\label{res}

%真正表1
\begin{table*}[htbp]
\centering

\caption{
Forecasting results with representative time-series and spatio-temporal baselines.
The best and second-best results are highlighted in \textbf{bold} and
\underline{underline}, respectively.
}
\label{tab:water_quality_long_horizon3}

\small
\setlength{\tabcolsep}{1.8pt}
\renewcommand{\arraystretch}{1.2}

\begin{adjustbox}{max width=\textwidth}

\begin{tabular}{
@{}cc|cc|cc|cc|cc|cc|cc|cc|cc
@{}}

\toprule

\multicolumn{2}{c|}{}
&
\multicolumn{8}{c|}{\textbf{Time-series Models}}
&
\multicolumn{6}{c|}{\textbf{Spatio-temporal Models}}
&
\multicolumn{2}{c}{\textbf{Ours}}
\\

\cmidrule(lr){3-10}
\cmidrule(lr){11-16}
\cmidrule(lr){17-18}

\multicolumn{2}{c|}{\textbf{Model}}
&
\multicolumn{2}{c|}{\textbf{TimeKAN}}
&
\multicolumn{2}{c|}{\textbf{FilterTS}}
&
\multicolumn{2}{c|}{\textbf{GTR}}
&
\multicolumn{2}{c|}{\textbf{MixLinear}}
&
\multicolumn{2}{c|}{\textbf{HyperD}}
&
\multicolumn{2}{c|}{\textbf{RSTIB-MLP}}
&
\multicolumn{2}{c|}{\textbf{ST-SSDL}}
&
\multicolumn{2}{c}{\textbf{RaiNet}}
\\

\cmidrule(lr){3-4}
\cmidrule(lr){5-6}
\cmidrule(lr){7-8}
\cmidrule(lr){9-10}
\cmidrule(lr){11-12}
\cmidrule(lr){13-14}
\cmidrule(lr){15-16}
\cmidrule(lr){17-18}

\multicolumn{2}{c|}{\textbf{Metric}}
&
\textbf{MSE} & \textbf{MAE}
&
\textbf{MSE} & \textbf{MAE}
&
\textbf{MSE} & \textbf{MAE}
&
\textbf{MSE} & \textbf{MAE}
&
\textbf{MSE} & \textbf{MAE}
&
\textbf{MSE} & \textbf{MAE}
&
\textbf{MSE} & \textbf{MAE}
&
\textbf{MSE} & \textbf{MAE}
\\

\midrule

\multirow{5}{*}{
\rotatebox[origin=c]{90}{\textbf{\texttt{170900}}}
}
& 24
& \underline{0.2628} & \underline{0.1667}
& 0.2692 & 0.1722
& 0.2763 & 0.1726
& 0.3790 & 0.2324
& 0.3786 & 0.1909
& 0.2960 & 0.1871
& 0.2792 & 0.2107
& \textbf{0.2185} & \textbf{0.1560} \\

& 48
& \underline{0.3687} & 0.2137
& 0.3697 & \underline{0.2131}
& 0.3757 & 0.2150
& 0.4640 & 0.2704
& 0.4554 & 0.2290
& 0.3958 & 0.2227
& 0.3739 & 0.2431
& \textbf{0.3162} & \textbf{0.2001} \\

& 96
& 0.4778 & \underline{0.2623}
& \underline{0.4771} & 0.2626
& 0.4834 & 0.2648
& 0.5739 & 0.3208
& 0.5518 & 0.2729
& 0.4994 & 0.2676
& 0.5410 & 0.3177
& \textbf{0.4096} & \textbf{0.2604} \\

& 144
& 0.5494 & 0.2998
& \underline{0.5433} & 0.2946
& 0.5458 & 0.2963
& 0.6449 & 0.3530
& 0.6081 & 0.3009
& 0.5538 & \underline{0.2895}
& 0.6011 & 0.3261
& \textbf{0.4904} & \textbf{0.2840} \\

& 192
& 0.6053 & 0.3204
& 0.5974 & 0.3199
& 0.5975 & 0.3207
& 0.7047 & 0.3786
& 0.6442 & 0.3183
& \underline{0.5937} & \textbf{0.3061}
& 0.6603 & 0.3691
& \textbf{0.5283} & \underline{0.3102} \\

\midrule

\multirow{5}{*}{
\rotatebox[origin=c]{90}{\textbf{\texttt{071200}}}
}
& 24
& \underline{0.1084} & 0.1905
& 0.1105 & \underline{0.1885}
& 0.1162 & 0.1967
& 0.2458 & 0.2848
& 0.6530 & 0.4605
& 0.1630 & 0.2555
& 0.1516 & 0.2535
& \textbf{0.0872} & \textbf{0.1747} \\

& 48
& 0.2425 & 0.2805
& \underline{0.2421} & \underline{0.2723}
& 0.2529 & 0.2851
& 0.3506 & 0.3429
& 0.6092 & 0.4694
& 0.2822 & 0.3126
& 0.5788 & 0.5108
& \textbf{0.1707} & \textbf{0.2414} \\

& 96
& \underline{0.4412} & 0.3802
& 0.4473 & \underline{0.3784}
& 0.4591 & 0.3894
& 0.5818 & 0.4424
& 0.7663 & 0.5291
& 0.4613 & 0.3912
& 0.7145 & 0.5240
& \textbf{0.3225} & \textbf{0.3344} \\

& 144
& 0.5820 & \underline{0.4450}
& 0.5952 & 0.4480
& 0.6019 & 0.4572
& 0.7111 & 0.5014
& 0.8402 & 0.5601
& \underline{0.5813} & 0.4499
& 1.1075 & 0.7135
& \textbf{0.4504} & \textbf{0.4010} \\

& 192
& 0.6771 & \underline{0.4918}
& 0.6879 & 0.4944
& 0.7010 & 0.5041
& 0.8131 & 0.5463
& 0.8453 & 0.5759
& \underline{0.6597} & 0.4930
& 1.0558 & 0.7019
& \textbf{0.5415} & \textbf{0.4443} \\

\midrule

\multirow{5}{*}{
\rotatebox[origin=c]{90}{\textbf{\texttt{051201}}}
}
& 24
& \underline{0.0074} & \underline{0.0328}
& 0.0075 & 0.0337
& 0.0081 & 0.0353
& 0.0194 & 0.0599
& 1.3482 & 0.7703
& 0.7583 & 0.6066
& 0.0217 & 0.0821
& \textbf{0.0072} & \textbf{0.0318} \\

& 48
& \underline{0.0182} & \underline{0.0530}
& 0.0185 & 0.0543
& 0.0196 & 0.0564
& 0.0288 & 0.0734
& 1.3237 & 0.7886
& 0.7933 & 0.6167
& 0.0319 & 0.1066
& \textbf{0.0166} & \textbf{0.0518} \\

& 96
& \underline{0.0462} & \underline{0.0889}
& 0.0468 & 0.0903
& 0.0491 & 0.0938
& 0.0609 & 0.1089
& 1.3572 & 0.8016
& 0.7660 & 0.6449
& 0.0756 & 0.1626
& \textbf{0.0414} & \textbf{0.0867} \\

& 144
& \underline{0.0785} & \underline{0.1214}
& 0.0789 & 0.1230
& 0.0807 & 0.1242
& 0.0963 & 0.1415
& 1.3682 & 0.8210
& 0.8104 & 0.6486
& 0.1314 & 0.2041
& \textbf{0.0680} & \textbf{0.1171} \\

& 192
& 0.1129 & 0.1524
& \underline{0.1113} & \underline{0.1505}
& 0.1151 & 0.1536
& 0.1378 & 0.1763
& 1.3842 & 0.8280
& 0.9683 & 0.7014
& 0.1491 & 0.2120
& \textbf{0.0992} & \textbf{0.1455} \\

\bottomrule
\end{tabular}

\end{adjustbox}
\end{table*}

%真正表2
\begin{table*}[htbp]
\centering

\caption{
Forecasting results with water-quality-specific and diffusion-based models.
The best and second-best results are highlighted in \textbf{bold} and
\underline{underline}, respectively.
}

\label{tab:water_quality_long_horizon4}

\small
\setlength{\tabcolsep}{1.8pt}
\renewcommand{\arraystretch}{1.2}

\begin{adjustbox}{max width=\textwidth}
\begin{tabular}{
@{}cc|cc|cc|cc|cc|cc|cc|cc|cc
@{}}

\toprule

\multicolumn{2}{c|}{}
&
\multicolumn{10}{c|}{\textbf{Water-quality-specific Models}}
&
\multicolumn{4}{c|}{\textbf{Diffusion-based Models}}
&
\multicolumn{2}{c}{\textbf{Ours}}
\\

\cmidrule(lr){3-12}
\cmidrule(lr){13-16}
\cmidrule(lr){17-18}

\multicolumn{2}{c|}{\textbf{Model}}
&
\multicolumn{2}{c|}{\textbf{PIWON}}
&
\multicolumn{2}{c|}{\textbf{D-TGCN}}
&
\multicolumn{2}{c|}{\textbf{CMLIP}}
&
\multicolumn{2}{c|}{\textbf{RGCN}}
&
\multicolumn{2}{c|}{\textbf{XGBR}}
&
\multicolumn{2}{c|}{\textbf{SimDiff}}
&
\multicolumn{2}{c|}{\textbf{MA-TSD}}
&
\multicolumn{2}{c}{\textbf{RaiNet}}
\\

\cmidrule(lr){3-4}
\cmidrule(lr){5-6}
\cmidrule(lr){7-8}
\cmidrule(lr){9-10}
\cmidrule(lr){11-12}
\cmidrule(lr){13-14}
\cmidrule(lr){15-16}
\cmidrule(lr){17-18}

\multicolumn{2}{c|}{\textbf{Metric}}
&
\textbf{MSE} & \textbf{MAE}
&
\textbf{MSE} & \textbf{MAE}
&
\textbf{MSE} & \textbf{MAE}
&
\textbf{MSE} & \textbf{MAE}
&
\textbf{MSE} & \textbf{MAE}
&
\textbf{MSE} & \textbf{MAE}
&
\textbf{MSE} & \textbf{MAE}
&
\textbf{MSE} & \textbf{MAE}
\\

\midrule

\multirow{5}{*}{
\rotatebox[origin=c]{90}{\textbf{\texttt{170900}}}
}
& 24
& \underline{0.2523} & 0.2013
& 0.2719 & 0.2111
& 0.2796 & 0.1794
& 0.2976 & 0.2113
& 0.3924 & 0.1913
& 0.2773 & \underline{0.1641}
& 0.2843 & 0.2023
& \textbf{0.2185} & \textbf{0.1560}
\\

& 48
& 0.3960 & 0.2634
& \underline{0.3658} & 0.2568
& 0.3875 & 0.2247
& 0.3897 & 0.2617
& 0.5593 & 0.2467
& 0.3982 & \underline{0.2142}
& 0.3877 & 0.2582
& \textbf{0.3162} & \textbf{0.2001}
\\

& 96
& 0.5303 & 0.3326
& \underline{0.4633} & 0.3072
& 0.4943 & 0.2723
& 0.4795 & 0.3179
& 0.8964 & 0.3228
& 0.5062 & \underline{0.2643}
& 0.5011 & 0.3314
& \textbf{0.4096} & \textbf{0.2604}
\\

& 144
& 0.6787 & 0.4255
& \underline{0.5201} & 0.3449
& 0.5500 & \underline{0.2992}
& 0.5312 & 0.3496
& 1.2130 & 0.3769
& 0.5781 & 0.2993
& 0.5739 & 0.3750
& \textbf{0.4904} & \textbf{0.2840}
\\

& 192
& 0.9224 & 0.5161
& \underline{0.5618} & 0.3661
& 0.6139 & 0.3338
& 0.5669 & 0.3673
& 1.4016 & 0.4131
& 0.6387 & \underline{0.3250}
& 0.6263 & 0.4098
& \textbf{0.5283} & \textbf{0.3102}
\\

\midrule

\multirow{5}{*}{
\rotatebox[origin=c]{90}{\textbf{\texttt{071200}}}
}
& 24
& \underline{0.0961} & \underline{0.1936}
& 0.1166 & 0.2043
& 0.1272 & 0.2011
& 0.1784 & 0.2460
& 0.1227 & 0.1999
& 0.1387 & 0.2068
& 0.1280 & 0.2168
& \textbf{0.0872} & \textbf{0.1747}
\\

& 48
& \underline{0.1868} & \underline{0.2676}
& 0.2278 & 0.2864
& 0.2631 & 0.2840
& 0.2855 & 0.3049
& 0.2514 & 0.2810
& 0.2544 & 0.2765
& 0.2643 & 0.3105
& \textbf{0.1707} & \textbf{0.2414}
\\

& 96
& \underline{0.4115} & \underline{0.3869}
& 0.4347 & 0.4086
& 0.4734 & 0.3919
& 0.4476 & 0.3912
& 0.4779 & 0.3951
& 0.4708 & 0.3911
& 0.4824 & 0.4257
& \textbf{0.3225} & \textbf{0.3344}
\\

& 144
& \underline{0.4967} & \underline{0.4261}
& 0.5533 & 0.4683
& 0.6243 & 0.4640
& 0.5665 & 0.4565
& 0.6684 & 0.4805
& 0.6278 & 0.4680
& 0.6267 & 0.4962
& \textbf{0.4504} & \textbf{0.4010}
\\

& 192
& \underline{0.5877} & \underline{0.4698}
& 0.6435 & 0.5066
& 0.7240 & 0.5107
& 0.6450 & 0.4971
& 0.8040 & 0.5385
& 0.7305 & 0.5194
& 0.7246 & 0.5501
& \textbf{0.5415} & \textbf{0.4443}
\\

\midrule

\multirow{5}{*}{
\rotatebox[origin=c]{90}{\textbf{\texttt{051201}}}
}
& 24
& 0.0090 & 0.0461
& 0.0093 & 0.0448
& \underline{0.0081} & \underline{0.0343}
& 0.0980 & 0.1883
& 0.0121 & 0.0551
& 0.0098 & 0.0401
& 0.0081 & 0.0367
& \textbf{0.0072} & \textbf{0.0318}
\\

& 48
& 0.0235 & 0.0797
& 0.0217 & 0.0698
& 0.0200 & 0.0561
& 0.1193 & 0.2011
& 0.0251 & 0.0754
& 0.0197 & \underline{0.0553}
& \underline{0.0197} & 0.0609
& \textbf{0.0166} & \textbf{0.0518}
\\

& 96
& 0.0555 & 0.1233
& 0.0569 & 0.1180
& 0.0516 & 0.0951
& 0.1394 & 0.2101
& 0.0565 & 0.1102
& 0.0530 & \underline{0.0930}
& \underline{0.0482} & 0.1009
& \textbf{0.0414} & \textbf{0.0867}
\\

& 144
& 0.1031 & 0.1763
& 0.0911 & 0.1549
& 0.0881 & 0.1287
& 0.1718 & 0.2461
& 0.0913 & 0.1413
& 0.0930 & 0.1335
& \underline{0.0797} & \underline{0.1287}
& \textbf{0.0680} & \textbf{0.1171}
\\

& 192
& 0.1316 & 0.1981
& 0.1241 & 0.1884
& 0.1206 & \underline{0.1558}
& 0.2055 & 0.2689
& 0.1272 & 0.1702
& 0.1260 & 0.1593
& \underline{0.1130} & 0.1642
& \textbf{0.0992} & \textbf{0.1455}
\\

\bottomrule
\end{tabular}
\end{adjustbox}

\end{table*}

Tables~\ref{tab:water_quality_long_horizon3} and
\ref{tab:water_quality_long_horizon4} report the complete results over five
forecasting horizons. RaiNet achieves the lowest MSE in all 15 settings and
the lowest MAE in 14 of them. On 170900, it reduces MSE by 5.7-13.6\%
relative to the strongest baseline across horizons. The sole exception in
MAE occurs at 192 steps, where RSTIB-MLP obtains 0.3061 compared with
RaiNet's 0.3102. On 071200, RaiNet maintains clear advantages at every
horizon, including a 21.6\% MSE reduction at 96 steps. On the finer-resolution
051201 dataset, it consistently improves over strong time-series,
water-quality-specific, and diffusion-based models, despite the smaller
absolute errors and consequently narrower margins.

The spatiotemporal results reveal an important mismatch between generic
cross-station modeling and basin-level water-quality forecasting.
Monitoring stations are sparsely distributed, while their observations are
affected by heterogeneous local environments and asynchronous responses.
Consequently, directly propagating station features through a predefined
spatial structure may introduce variance without providing consistently
useful predictive information. This is reflected by the unstable performance
of HyperD, RSTIB-MLP, and ST-SSDL: they remain competitive in a few 170900
settings, but fall behind on 071200, while HyperD and RSTIB-MLP deteriorate
substantially on 051201. Thus, their limitations arise not from spatial
modeling itself, but from treating cross-station dependence as a uniformly
reliable forecasting signal.

RaiNet avoids this failure mode by modeling water-quality dynamics
station-wise and introducing spatial information asymmetrically through
station-related rainfall events and basin context. Its consistent performance
across different targets, temporal resolutions, station counts, and basin
configurations demonstrates that conditional environmental fusion generalizes
more reliably than direct cross-station propagation for these datasets.
Table~\ref{tab:water_quality_long_horizon4} further shows that this advantage
extends beyond generic forecasting models: RaiNet also consistently
outperforms water-quality-specific and diffusion-based alternatives,
indicating that neither specialized spatial structures nor generative
forecasting alone adequately resolves heterogeneous rainfall and water-quality
dynamics.

\begin{wrapfigure}{r}{0.58\columnwidth}
    \vspace{-1.2em}
    \centering
    \includegraphics[width=\linewidth]{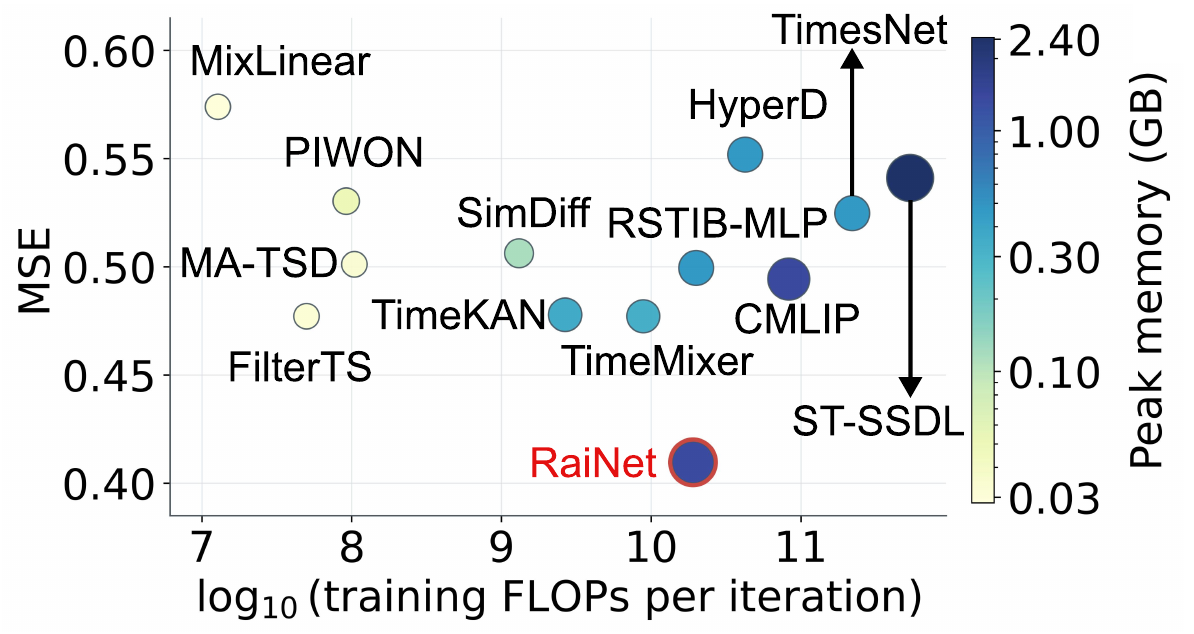}
    \caption{Accuracy-efficiency trade-offs on 170900 with an input and
prediction length of 96. Bubble size and color denote peak allocated GPU
memory.}
    \label{fig:complexity_tradeoff}
    %\vspace{-1.5em}
\end{wrapfigure}

Figure~\ref{fig:complexity_tradeoff} compares forecasting accuracy,
computational cost, and peak GPU memory under a unified setting. All models
are evaluated on basin 170900 using an input length of 96, a prediction
length of 96, and a batch size of 64 on a single NVIDIA GeForce RTX 5080.
Training FLOPs are measured for one complete forward-and-backward iteration,
while memory denotes the peak GPU memory allocated during execution. Despite incorporating an additional gridded-rainfall modality, RaiNet requires
only 19.03 GFLOPs per training iteration and 1.32 GB of peak allocated memory.
Its computational cost is 55.2\% lower than HyperD, 77.1\% lower than CMLIP,
91.4\% lower than TimesNet, and 96.4\% lower than ST-SSDL, while remaining
slightly below RSTIB-MLP. Although several lightweight temporal baselines
require fewer operations, they produce consistently higher forecasting
errors. RaiNet therefore occupies a favorable region of the
accuracy--efficiency space, achieving the lowest MSE without the excessive
computation required by several complex forecasting models. The modest absolute memory requirement also leaves substantial capacity for
larger batches or concurrent forecasting tasks on a single RTX 5080.
Combined with its station-wise prediction structure and efficient conditional
integration of rainfall information, these results indicate that RaiNet is
practical for scheduled basin-scale forecasting services without requiring
multi-GPU infrastructure.

\subsection{DETAILED ABLATION STUDIES}
\label{abala}

\paragraph{Cumulative construction.}
Starting from the temporal-only model with moving-average decomposition
(\textit{T-only (MA)}), replacing the fixed moving average with
\texttt{LocTrend} already reduces both MSE and MAE. This improvement indicates
that modeling the structural background through locally matched patterns is
more effective than imposing a fixed smoothing rule. Introducing multiscale
modeling further improves prediction, suggesting that water-quality dynamics
cannot be sufficiently characterized at a single temporal resolution.

The incorporation of the rainfall event encoder $\mathcal{E}_I$ produces
another clear reduction in error. This result supports the use of
station-oriented rainfall events rather than relying solely on the historical
water-quality sequence. Adding \texttt{XGateFusion} yields a further substantial
gain, showing that rainfall information is more useful when its contribution is
adaptively conditioned on the temporal baseline instead of being injected
uniformly. The joint predictor $P_C$ subsequently improves the result by
integrating the enhanced temporal representation with basin rainfall context,
and the final directed topology $\mathcal{G}$ provides the last improvement by
incorporating inter-station dependencies along the river network. Importantly, the
improvement remains monotonic throughout the construction, indicating that the
modules are complementary.

\paragraph{Standalone architectural variants.}
The standalone variants provide a more targeted diagnosis of the proposed
design. Setting $L=1$ removes cross-scale modeling while retaining the remaining
components. Its degradation confirms that the gain of RaiNet does not arise
only from increasing model capacity, but also from explicitly representing
water-quality dynamics at multiple temporal resolutions. Replacing \texttt{LocTrend} with a moving-average decomposition
(\texttt{LocTrend}$\rightarrow$\texttt{MA}) causes a clear performance drop.
This comparison directly isolates the decomposition strategy and shows that a
fixed local smoother cannot fully substitute for the adaptive structural
dictionary used by \texttt{LocTrend}. In particular, the result is consistent
with the motivation that locally evolving backgrounds need not follow a fixed
seasonal or smoothing pattern.

Removing the joint predictor $P_C$ also increases both errors, demonstrating
that the scale-wise rainfall-conditioned predictions still benefit from a final
joint integration of temporal, rainfall, and spatial information. Similarly,
removing $\mathcal{E}_I$ leads to a noticeable degradation, confirming that the
basin-relative rainfall representation provides useful information beyond the
water-quality history itself. The deterioration after removing $\mathcal{G}$
further indicates that the directed river topology contributes complementary
spatial information rather than merely duplicating temporal correlations.

\paragraph{XGateFusion diagnostics.}
Replacing \texttt{XGateFusion} with a residual gate results in higher errors (\texttt{XGateFusion}$\rightarrow$\texttt{RG}),
showing that a simple global residual weighting is insufficient to capture the
conditional effect of rainfall. The proposed mechanism instead compares the
temporal baseline with its rainfall-conditioned counterpart and constructs
station- and horizon-dependent corrections from their relative-lag interaction. The remaining variants further separate the two levels of gating used by
\texttt{XGateFusion}. Removing the dynamic update
$-\Delta\mathbf{G}^{(l)}$ retains only the scale-level base contribution,
whereas removing $-g^{(l)}$ forces the fusion to rely only on the dynamic
interaction-derived correction. Both variants perform worse than the complete
model, indicating that the two terms play complementary roles: $g^{(l)}$
captures the overall reliability of rainfall information at each scale, while
$\Delta\mathbf{G}^{(l)}$ adjusts this contribution according to station and
forecast-horizon-specific interactions.

Finally, removing the explicit background-deviation decomposition
$-(\mathbf{B}^{(l)},\mathbf{D}^{(l)})$ also degrades performance. This result
shows that the benefit of \texttt{LocTrend} is not limited to producing an
alternative temporal representation; preserving the structural background and
dynamic deviation as distinct components is itself useful for subsequent
cross-scale modeling. Together, these standalone tests support the main design
principle of RaiNet: locally evolving temporal structure, multiscale dynamics,
rainfall-conditioned corrections, and river-network dependencies provide
different and complementary sources of predictive information.

\subsection{Qualitative Analysis}
\label{qa}
\begin{figure}[htbp]
    \centering
    \includegraphics[width=1\linewidth]{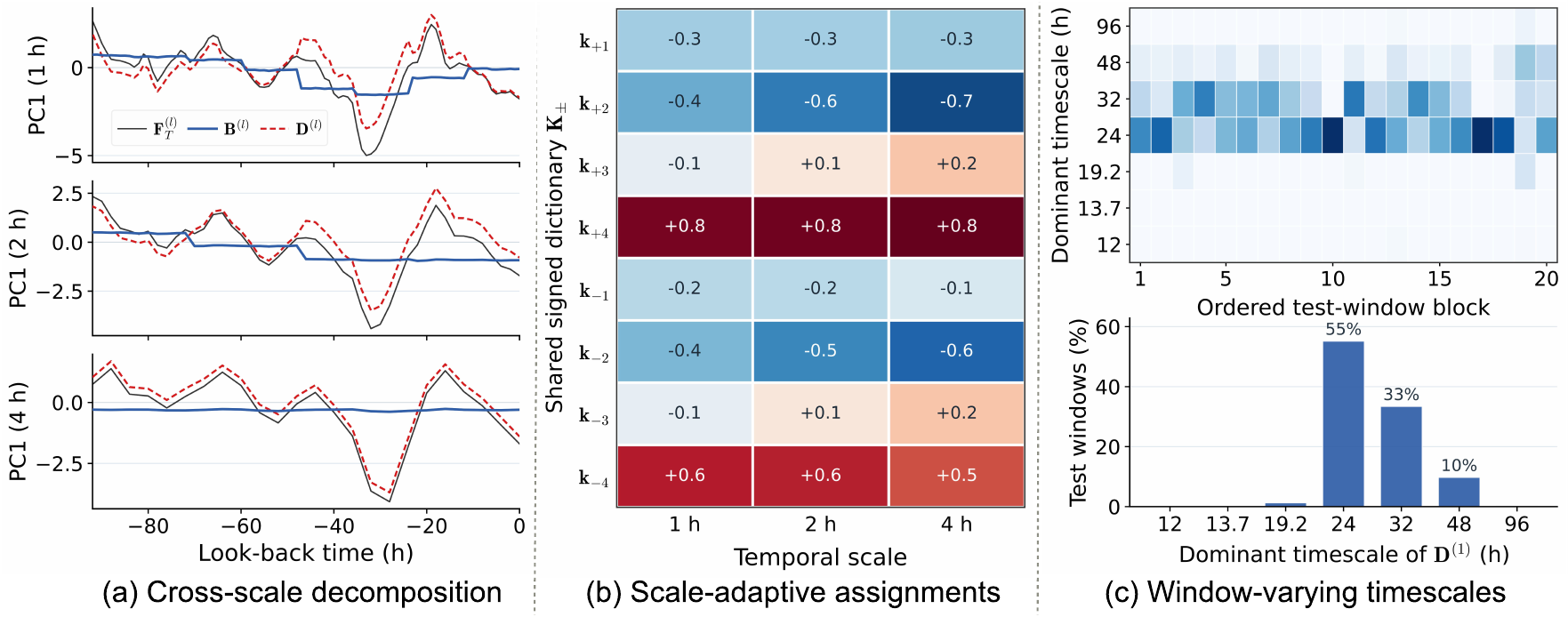}
    \caption{\textbf{Analysis of \texttt{LocTrend}.}
(a) At different temporal scales, \texttt{LocTrend} separates the input
$\mathbf{F}_T^{(l)}$ into a locally evolving structural background
$\mathbf{B}^{(l)}$ and an exact dynamic deviation $\mathbf{D}^{(l)}$.
(b) The three scales exhibit different soft-assignment patterns over the
shared signed dictionary $\mathbf{K}_{\pm}$.
(c) Post-hoc spectral analysis shows window-varying dominant timescales in $\mathbf{D}^{(1)}$. The heatmap and bars show their block-wise and overall distributions.}
    \label{fig:loctrend_analysis}
\end{figure}

We further examine whether \texttt{LocTrend} realizes the intended
background-deviation decomposition while remaining adaptive to temporal
resolution and local dynamics. Figure~\ref{fig:loctrend_analysis}(a) first
visualizes the decomposition at the 1-, 2-, and 4-hour scales. At each scale,
\texttt{LocTrend} satisfies
$\mathbf{F}_{T}^{(l)}=\mathbf{B}^{(l)}+\mathbf{D}^{(l)}$,
where $\mathbf{B}^{(l)}$ captures the locally evolving structural background
and $\mathbf{D}^{(l)}$ is its exact complementary deviation. The resulting
background becomes progressively smoother as the temporal resolution becomes
coarser, while the deviation retains the remaining local variations. This
scale-dependent behavior indicates that \texttt{LocTrend} does not impose an
identical decomposition pattern across resolutions.

Figure~\ref{fig:loctrend_analysis}(b) further reveals the mechanism underlying
this adaptation. All decomposed scales share the same signed structural
dictionary $\mathbf{K}_{\pm}$, but their soft-assignment patterns differ
substantially. In particular, both the utilization of different atom pairs and
the preference between their positive and negative directions vary across
scales. Since the reconstructed background is determined by
$\widetilde{\mathbf{b}}_{i,t}^{(l)}
=
\sum_{r=1}^{2K}
a_{i,t,r}^{(l)}\mathbf{k}_{\pm,r},$ the scale dependence therefore arises from the assignment coefficients rather than from introducing separate dictionaries for
different resolutions. This shared-dictionary yet scale-adaptive reconstruction
allows common structural patterns to be reused while their contributions are
adjusted according to the temporal scale.

Finally, Figure~\ref{fig:loctrend_analysis}(c) examines whether the resulting
decomposition is tied to a fixed periodicity. Even at the same 1-hour
resolution, the dominant timescale of $\mathbf{D}^{(1)}$ varies considerably
across test windows, with prominent occurrences around 24, 32, and 48 hours.
These timescales are obtained only through post-hoc spectral analysis of the
decomposed deviation; they are neither predefined in \texttt{LocTrend} nor used
during dictionary fitting or forward reconstruction. Their variation across
windows therefore provides empirical evidence that the decomposition is not
anchored to a prescribed seasonal period. The three analyses show
that \texttt{LocTrend} achieves scale adaptation through data-dependent
structural assignment while accommodating locally varying temporal
organizations without assuming fixed periodicity.

% Figure~\ref{fig:rain_xgate_analysis} examines how rainfall information is transformed and used by RaiNet.
% As shown in Fig.~\ref{fig:rain_xgate_analysis}(a), comparable station-local rainfall can occur under substantially different basin-wide rainfall conditions. Accordingly, the rainfall event encoder produces different station-aligned event states, demonstrating that it captures basin-relative event semantics rather than absolute local intensity alone.

% To isolate the conditional behavior of XGateFusion, Fig.~\ref{fig:rain_xgate_analysis}(b) considers a separate pair of samples with the same rainfall-event representation but different water-quality histories. Their relative-lag profiles differ across both directions and temporal scales, showing that lag matching is conditioned on the current water-quality context rather than determined by rainfall alone. These lags represent statistical relative offsets, not causal propagation times.

% Figure~\ref{fig:rain_xgate_analysis}(c) further shows that the learned matching is translated into signed prediction corrections. The correction magnitude and direction vary across scales and stations, while several branches remain close to zero. This confirms that XGateFusion uses rainfall selectively instead of applying a uniform correction to the water-quality baseline.

\end{document}

%% file: math_commands.tex
\usepackage{amsmath,amsfonts,bm}

\def\eqref#1{equation~\ref{#1}}
\def\1{\bm{1}}

\DeclareMathAlphabet{\mathsfit}{\encodingdefault}{\sfdefault}{m}{sl}
\SetMathAlphabet{\mathsfit}{bold}{\encodingdefault}{\sfdefault}{bx}{n}

%% file: iclr2027_conference.bib
@inproceedings{
TimeMixer,
title={TimeMixer: Decomposable Multiscale Mixing for Time Series Forecasting},
author={Shiyu Wang and Haixu Wu and Xiaoming Shi and Tengge Hu and Huakun Luo and Lintao Ma and James Y. Zhang and JUN ZHOU},
booktitle={The Twelfth International Conference on Learning Representations},
year={2024},
url={https://openreview.net/forum?id=7oLshfEIC2}
}

@inproceedings{
Timesnet,
title={TimesNet: Temporal 2D-Variation Modeling for General Time Series Analysis},
author={Haixu Wu and Tengge Hu and Yong Liu and Hang Zhou and Jianmin Wang and Mingsheng Long},
booktitle={The Eleventh International Conference on Learning Representations },
year={2023},
url={https://openreview.net/forum?id=ju_Uqw384Oq}
}

@inproceedings{FEDformer,
  title={FEDformer: Frequency Enhanced Decomposed Transformer for Long-term Series Forecasting},
  author={Zhou, Tian and Ma, Ziqing and Wen, Qingsong and Wang, Xue and Sun, Liang and Jin, Rong},
  booktitle={Proceedings of the 39th International Conference on Machine Learning (ICML)},
  year={2022},
  note={arXiv:2201.12740},
  url={https://arxiv.org/abs/2201.12740}
}

@article{TimeKAN,
  title={TimeKAN: KAN-based Frequency Decomposition Learning Architecture for Long-term Time Series Forecasting},
  author={Huang, Songtao and Zhao, Zhen and Li, Can and Bai, Lei},
  journal={arXiv preprint arXiv:2502.06910},
  year={2025},
  url={https://arxiv.org/abs/2502.06910}
}

@inproceedings{FilterTS,
  title={FilterTS: Comprehensive Frequency Filtering for Multivariate Time Series Forecasting},
  author={Wang, Yulong and Liu, Yushuo and Duan, Xiaoyi and Wang, Kai},
  booktitle={Proceedings of the AAAI Conference on Artificial Intelligence},
  volume={39},
  pages={35438--35445},
  year={2025},
  doi={10.1609/aaai.v39i20.35438}
}

@INPROCEEDINGS{SARIMA,
  author={Sathya, R. and Steve, Leander P and Narayan, Nitish and Upadhye, Abhishek and Aravindharaj, M.},
  booktitle={2023 IEEE International Students' Conference on Electrical, Electronics and Computer Science (SCEECS)}, 
  title={Covid Wave Prediction using SARIMA Machine Learning Algorithm}, 
  year={2023},
  volume={},
  number={},
  pages={1-5},
  doi={10.1109/SCEECS57921.2023.10062979}}

@ARTICLE{STL,
  author={Yan, Duojin and Yang, Chunjie and Sun, Shuming and Lou, Siwei and Kong, Liyuan and Zhang, Yuyan},
  journal={IEEE Transactions on Instrumentation and Measurement}, 
  title={One-Sided Relational Autoencoder With Seasonal-Trend Decomposition to Extract Process Correlations for Molten Iron Quality Prediction}, 
  year={2024},
  volume={73},
  number={},
  pages={1-13},
  doi={10.1109/TIM.2024.3353844}}

@INPROCEEDINGS{MSE,
  author={Richman, Steven H. and Schwartz, Mischa},
  booktitle={1971 IEEE Conference on Decision and Control}, 
  title={Dynamic programming training period for an MSE adaptive equalizer}, 
  year={1971},
  volume={},
  number={},
  pages={262-263},
  doi={10.1109/CDC.1971.270994}}

@INPROCEEDINGS{MAE,
  author={Menez, J. and Boeri, F. and Esteban, D.},
  booktitle={ICASSP '79. IEEE International Conference on Acoustics, Speech, and Signal Processing}, 
  title={Optimum quantizer algorithm for real-time block quantizing}, 
  year={1979},
  volume={4},
  number={},
  pages={980-984},
  doi={10.1109/ICASSP.1979.1170824}}

@article{CDA,
  title={Mutualformer: Multi-modal representation learning via cross-diffusion attention},
  author={Wang, Xixi and Wang, Xiao and Jiang, Bo and Tang, Jin and Luo, Bin},
  journal={International Journal of Computer Vision},
  volume={132},
  number={9},
  pages={3867--3888},
  year={2024},
  publisher={Springer}
}

@INPROCEEDINGS{MBT,
  author={Papillon, Olivier and Goubran, Rafik and Green, James and Larivière-Chartier, Julien and Higginson, Caitlin and Knoefel, Frank and Robillard, Rébecca},
  booktitle={2025 IEEE Medical Measurements and Applications (MeMeA)}, 
  title={Sleep Stage Classification using Multimodal Embedding Fusion from Electrooculography and Pressure-Sensitive Mats}, 
  year={2025},
  volume={},
  number={},
  pages={1-6},
  doi={10.1109/MeMeA65319.2025.11068039}}

@ARTICLE{dataset,
  author={Bi, Jing and Wu, Xiangxi and Yuan, Haitao and Wang, Ziqi and Wei, Daming and Wu, Renren and Zhang, Jia and Qiao, Junfei and Buyya, Rajkumar},
  journal={IEEE Internet of Things Journal}, 
  title={STMF: A Spatiotemporal Multimodal Fusion Model for Long-Term Water Quality Forecasting}, 
  year={2025},
  volume={12},
  number={18},
  pages={37146-37159},
  doi={10.1109/JIOT.2025.3581282}}

@ARTICLE{ma,
  author={Pappan, Balakrishnan and Sankaran, Durairaj and Selvaraj, Sathiya},
  journal={IEEE Canadian Journal of Electrical and Computer Engineering}, 
  title={Advanced Prediction and Coated Solar Panel Performance Improvement Using Combined Long Short-Term Memory (LSTM) Architecture–Autoregressive Moving Average (ARMA) Technique}, 
  year={2025},
  volume={48},
  number={4},
  pages={313-324},
  doi={10.1109/ICJECE.2025.3592010}}

@incollection{ARIMAnew,
  title={ARIMA models},
  author={Shumway, Robert H and Stoffer, David S},
  booktitle={Time series analysis and its applications: with R examples},
  pages={75--163},
  year={2017},
  publisher={Springer}
}

@article{HW,
  title={The Holt-winters forecasting procedure},
  author={Chatfield, Chris},
  journal={Journal of the Royal Statistical Society: Series C (Applied Statistics)},
  volume={27},
  number={3},
  pages={264--279},
  year={1978},
  publisher={Wiley Online Library}
}

@inproceedings{VT,
  title={Time-VLM: Exploring Multimodal Vision-Language Models for Augmented Time Series Forecasting},
  author={Siru, Zhong and Weilin, Ruan and Jin, Ming and Huan, Li and Qingsong, Wen and Yuxuan, Liang},
  booktitle={Forty-Second International Conference on Machine Learning (ICML 2025)},
  year={2025},
  organization={Proceedings of Machine Learning Research}
}

@article{ho2025graph,
  title={Graph anomaly detection in time series: A survey},
  author={Ho, Thi Kieu Khanh and Karami, Ali and Armanfard, Narges},
  journal={IEEE Transactions on Pattern Analysis and Machine Intelligence},
  year={2025},
  publisher={IEEE}
}

@inproceedings{graph111,
  title={Motif-aware Graph Neural Networks for Networked Time Series Imputation},
  author={Ahmed, Nourhan and Yalavarthi, Vijaya Krishna and Schmidt-Thieme, Lars},
  booktitle={Proceedings of the AAAI Conference on Artificial Intelligence},
  volume={39},
  pages={11409--11417},
  year={2025}
}

@article{krysanova2015advances,
  title={Advances in water resources assessment with SWAT—an overview},
  author={Krysanova, Valentina and White, Mike},
  journal={Hydrological Sciences Journal},
  volume={60},
  number={5},
  pages={771--783},
  year={2015},
  publisher={Taylor \& Francis}
}

@article{wool2020wasp,
  title={WASP 8: The next generation in the 50-year evolution of USEPA’s water quality model},
  author={Wool, Tim and Ambrose Jr, Robert B and Martin, James L and Comer, Alex},
  journal={Water},
  volume={12},
  number={5},
  pages={1398},
  year={2020},
  publisher={MDPI}
}

@article{zhang2012simulation,
  title={Simulation of water environmental capacity and pollution load reduction using QUAL2K for water environmental management},
  author={Zhang, Ruibin and Qian, Xin and Yuan, Xingcheng and Ye, Rui and Xia, Bisheng and Wang, Yulei},
  journal={International journal of environmental research and public health},
  volume={9},
  number={12},
  pages={4504--4521},
  year={2012},
  publisher={MDPI}
}

@article{chen2024coupled,
  title={A coupled model to improve river water quality prediction towards addressing non-stationarity and data limitation},
  author={Chen, Shengyue and Huang, Jinliang and Wang, Peng and Tang, Xi and Zhang, Zhenyu},
  journal={Water Research},
  volume={248},
  pages={120895},
  year={2024},
  publisher={Elsevier}
}

@article{wan2025temporal,
  title={Temporal and spatial feature extraction using graph neural networks for multi-point water quality prediction in river network areas},
  author={Wan, Hang and Xiang, Long and Cai, Yanpeng and Xie, Yulei and Xu, Rui},
  journal={Water Research},
  volume={281},
  pages={123561},
  year={2025},
  publisher={Elsevier}
}

@article{bi2025long,
  title={Long-term water quality prediction with transformer-based spatial-temporal graph fusion},
  author={Bi, Jing and Wang, Ziqi and Yuan, Haitao and Wu, Xiangxi and Wu, Renren and Zhang, Jia and Zhou, MengChu},
  journal={IEEE Transactions on Automation Science and Engineering},
  volume={22},
  pages={11392--11404},
  year={2025},
  publisher={IEEE}
}

@article{zhang2023monitoring,
  title={Monitoring and spatial traceability of river water quality using Sentinel-2 satellite images},
  author={Zhang, Yingyin and He, Xianqiang and Lian, Gang and Bai, Yan and Yang, Ying and Gong, Fang and Wang, Difeng and Zhang, Zili and Li, Teng and Jin, Xuchen},
  journal={Science of The Total Environment},
  volume={894},
  pages={164862},
  year={2023},
  publisher={Elsevier}
}

@article{wang2026improving,
  title={Improving watershed-scale daily nutrient simulation using a process-model-informed graph attention network with multi-source data integration},
  author={Wang, Weichen and Liu, Guowangchen and Wang, Mingjing and Pan, Yan and Ma, Yukun and Yang, Lu and Sang, Jing and Shen, Zhenyao and Chen, Lei},
  journal={Water Research},
  pages={125532},
  year={2026},
  publisher={Elsevier}
}

@article{ortiz2023methodology,
  title={A methodology for integrating time-lagged rainfall and river flow data into machine learning models to improve prediction of quality parameters of raw water supplying a treatment plant},
  author={Ortiz-Lopez, Christian and Torres, Andres and Bouchard, Christian and Rodriguez, Manuel},
  journal={Journal of Hydroinformatics},
  volume={25},
  number={6},
  pages={2406--2426},
  year={2023},
  publisher={IWA Publishing}
}

@article{huang2024riverine,
  title={Which riverine water quality parameters can be predicted by meteorologically-driven deep learning?},
  author={Huang, Sheng and Wang, Yueling and Xia, Jun},
  journal={Science of The Total Environment},
  volume={946},
  pages={174357},
  year={2024},
  publisher={Elsevier}
}

@article{baek2020prediction,
  title={Prediction of water level and water quality using a CNN-LSTM combined deep learning approach},
  author={Baek, Sang-Soo and Pyo, Jongcheol and Chun, Jong Ahn},
  journal={Water},
  volume={12},
  number={12},
  pages={3399},
  year={2020},
  publisher={MDPI}
}

@article{bi2025hybrid,
  title={Hybrid water quality prediction with multimodal low-rank fusion and localized attention},
  author={Bi, Jing and Li, Yibo and Yuan, Haitao and Wang, Mengyuan and Wang, Ziqi and Zhang, Jia and Zhou, Mengchu},
  journal={IEEE Internet of Things Journal},
  volume={12},
  number={12},
  pages={21158--21169},
  year={2025},
  publisher={IEEE}
}

@article{wu2021autoformer,
  title={Autoformer: Decomposition transformers with auto-correlation for long-term series forecasting},
  author={Wu, Haixu and Xu, Jiehui and Wang, Jianmin and Long, Mingsheng},
  journal={Advances in neural information processing systems},
  volume={34},
  pages={22419--22430},
  year={2021}
}

@article{smith2024predictive,
  title={Predictive understanding of stream salinization in a developed watershed using machine learning},
  author={Smith, Jared D and Koenig, Lauren E and Sleckman, Margaux J and Appling, Alison P and Sadler, Jeffrey M and DePaul, Vincent T and Szabo, Zoltan},
  journal={Environmental Science \& Technology},
  volume={58},
  number={42},
  pages={18822},
  year={2024}
}

@article{jiang2024heterogeneous,
  title={Heterogeneous dynamic graph convolutional networks for enhanced spatiotemporal flood forecasting by remote sensing},
  author={Jiang, Jiange and Chen, Chen and Zhou, Yang and Berretti, Stefano and Liu, Lei and Pei, Qingqi and Zhou, Jianming and Wan, Shaohua},
  journal={IEEE Journal of Selected Topics in Applied Earth Observations and Remote Sensing},
  volume={17},
  pages={3108--3122},
  year={2024},
  publisher={IEEE}
}

@inproceedings{ding2026simdiff,
  title={SimDiff: Simpler Yet Better Diffusion Model for Time Series Point Forecasting},
  author={Ding, Hang and Wang, Xue and Zhou, Tian and Yao, Tao},
  booktitle={Proceedings of the AAAI Conference on Artificial Intelligence},
  volume={40},
  number={25},
  pages={20781--20789},
  year={2026}
}

@inproceedings{wang2025non,
  title={A non-isotropic time series diffusion model with moving average transitions},
  author={Wang, Chenxi and Yang, Linxiao and Wang, Zhixian and Sun, Liang and Wang, Yi},
  booktitle={Forty-second International Conference on Machine Learning},
  year={2025}
}

@article{ali2025near,
  title={Near real-time and next-day prediction for Escherichia coli (E. coli) concentrations in highly urbanized watersheds},
  author={Ali, Salou Moumouni and Imtiaz, Syed Usama and Mitra, Nasr Azadani and Nasrin, Alamdari},
  journal={Water Research},
  pages={125030},
  year={2025},
  publisher={Elsevier}
}

@article{tang2026non,
  title={Non-point source pollution prediction and dynamics simulation in urban runoff: a physics-informed neural network approach},
  author={Tang, Sijie and Jiang, Jiping and Wang, Shuo and Zheng, Yi and Savic, Dragan and Wang, Aijie},
  journal={Water Research},
  pages={126379},
  year={2026},
  publisher={Elsevier}
}

@inproceedings{
cao2026gtr,
title={Enhancing Multivariate Time Series Forecasting with Global Temporal Retrieval},
author={Fanpu Cao and Lu Dai and Jindong Han and Hui Xiong},
booktitle={The Fourteenth International Conference on Learning Representations},
year={2026},
url={https://openreview.net/forum?id=QUJBPSfyui}
}

@inproceedings{ma2026mixlinear,
  title={MixLinear: Extreme Low Resource Multivariate Time Series Forecasting with $0.1 K $ Parameters},
  author={Ma, Aitian and Luo, Dongsheng and Sha, Mo},
  booktitle={International Conference on Learning Representations},
  volume={2026},
  pages={77764--77789},
  year={2026}
}

@inproceedings{shao2026hyperd,
  title={Hyperd: Hybrid periodicity decoupling framework for traffic forecasting},
  author={Shao, Minlan and Zhang, Zijian and Wang, Yili and Dai, Yiwei and Shen, Xu and Wang, Xin},
  booktitle={Proceedings of the AAAI Conference on Artificial Intelligence},
  volume={40},
  number={18},
  pages={15689--15697},
  year={2026}
}

@inproceedings{chen2025information,
  title={Information bottleneck-guided mlps for robust spatial-temporal forecasting},
  author={Chen, Min and Pang, Guansong and Wang, Wenjun and Yan, Cheng},
  booktitle={Forty-second International Conference on Machine Learning},
  year={2025}
}

@article{gao2026different,
  title={How different from the past? spatio-temporal time series forecasting with self-supervised deviation learning},
  author={Gao, Haotian and Dong, Zheng and Yong, Jiawei and Fukushima, Shintaro and Taura, Kenjiro and Jiang, Renhe},
  journal={Advances in Neural Information Processing Systems},
  volume={38},
  pages={146647--146675},
  year={2026}
}

@article{zhang2026incorporating,
  title={Incorporating Flow Direction as a Physical Prior in Graph Networks: A Directed Temporal Graph Convolutional Network for Enhanced Water Quality Forecasting in Lake Basins},
  author={Zhang, Xuanran and He, Yufeng and Zhou, Zikuan and Li, Weihao and Tao, Lizhi and Chen, Jie and Zou, Haibo and Lin, Hui},
  journal={Transactions in GIS},
  volume={30},
  number={4},
  pages={e70309},
  year={2026},
  publisher={Wiley Online Library}
}
